\documentclass[compsoc,journal]{IEEEtran}
\usepackage{amsmath,amsfonts}
\usepackage{algorithmic}
\usepackage{algorithm}
\usepackage{array}
\usepackage[caption=false,font=normalsize,labelfont=sf,textfont=sf]{subfig}
\usepackage{textcomp}
\usepackage{stfloats}
\usepackage{url}
\usepackage{verbatim}
\usepackage{graphicx}
\usepackage{cite}
\usepackage{todonotes}

\usepackage{multirow}
\usepackage{booktabs}
\usepackage{hyperref}
\usepackage{color,soul}
\usepackage{colortbl}
\usepackage{lipsum}
\usepackage{mathrsfs}  
\usepackage{makecell}
\usepackage{threeparttable}
\usepackage[accsupp]{axessibility}
\usepackage{bm}
\definecolor{yellow}{rgb}{0.98,0.82,0.69}
\renewcommand{\algorithmicrequire}{\textbf{Input:}}
\graphicspath{{Figures/}}  

\begin{document}

\title{ReactFace: Online Multiple Appropriate Facial Reaction Generation in Dyadic Interactions}


\author{Cheng Luo,~\IEEEmembership{}
        Siyang Song$^*$,~\IEEEmembership{}
        Weicheng Xie$^*$,~\IEEEmembership{}
        Micol Spitale,~\IEEEmembership{}
        Zongyuan Ge,~\IEEEmembership{}
        Linlin Shen~\IEEEmembership{}
        and~Hatice Gunes~\IEEEmembership{}

\IEEEcompsocitemizethanks{\IEEEcompsocthanksitem Cheng Luo, Weicheng Xie and Linlin Shen are with Computer Vision Institute, College of Computer Science and Software Engineering, Shenzhen University, Shenzhen 518060, China. Weicheng Xie and Linlin Shen are also with Guangdong Provincial Key Laboratory of Intelligent Information Processing, Shenzhen University, Shenzhen 518060, China. Linlin Shen is also with the Department of Computer Science, University of Nottingham Ningbo China, Ningbo 315100, China.
\IEEEcompsocthanksitem Siyang Song is with Computer Sciences, University of Exeter, Exeter, EX4 4PY, United Kingdom.
\IEEEcompsocthanksitem Siyang Song, Micol Spitale and Hatice Gunes are with the Department of Computer Science and Technology, University of Cambridge, Cambridge, CB3 0FT, United Kingdom. E-mail: \{ss2796, ms2871, hg410\}@cam.ac.uk.
\IEEEcompsocthanksitem Cheng Luo and Zongyuan Ge are with the Airdoc-Monash Research Centre and Monash University Faculty of IT, 
Monash University, Melbourne, Australia. E-mail: 
\{cheng.luo, zongyuan.ge\}@monash.edu.

\IEEEcompsocthanksitem Corresponding Author: $^*$ Dr Siyang Song and Dr Weicheng Xie. E-mail: ss2796@cam.ac.uk and wcxie@szu.edu.cn.
}

}

\markboth{IEEE TRANSACTIONS ON VISUALIZATION AND COMPUTER GRAPHICS}%
{Shell \MakeLowercase{\textit{et al.}}: A Sample Article Using IEEEtran.cls for IEEE Journals}


\IEEEtitleabstractindextext{%
\begin{abstract}

In dyadic interaction, predicting the listener's facial reactions is challenging as different reactions could be appropriate in response to the same speaker's behaviour. Previous approaches predominantly treated this task as an interpolation or fitting problem, emphasizing deterministic outcomes but ignoring the diversity and uncertainty of human facial reactions. Furthermore, these methods often failed to model short-range and long-range dependencies within the interaction context, leading to issues in the synchrony and appropriateness of the generated facial reactions. To address these limitations, this paper reformulates the task as an extrapolation or prediction problem, and proposes an novel framework (called ReactFace) to generate multiple different but appropriate facial reactions from a speaker behaviour rather than merely replicating the corresponding listener facial behaviours.
Our ReactFace generates multiple different but appropriate photo-realistic human facial reactions by: (i) learning an appropriate facial reaction distribution representing multiple different but appropriate facial reactions; and (ii) synchronizing the generated facial reactions with the speaker verbal and non-verbal behaviours at each time stamp, resulting in realistic 2D facial reaction sequences. Experimental results demonstrate the effectiveness of our approach in generating multiple diverse, synchronized, and appropriate facial reactions from each speaker's behaviour. The quality of the generated facial reactions is intimately tied to the speaker's speech and facial expressions, achieved through our novel speaker-listener interaction modules.
Our code is made publicly available at \url{https://github.com/lingjivoo/ReactFace}.

\end{abstract}

\begin{IEEEkeywords}
Multiple Appropriate Facial Reaction Generation (MAFRG), Generative Model, Video Generation
\end{IEEEkeywords}}

\maketitle

\IEEEdisplaynontitleabstractindextext

%
\IEEEpeerreviewmaketitle

\section{Introduction}
\label{sec:introduction}

\begin{figure}[t!]
    \centering
    \includegraphics[width=1\columnwidth]{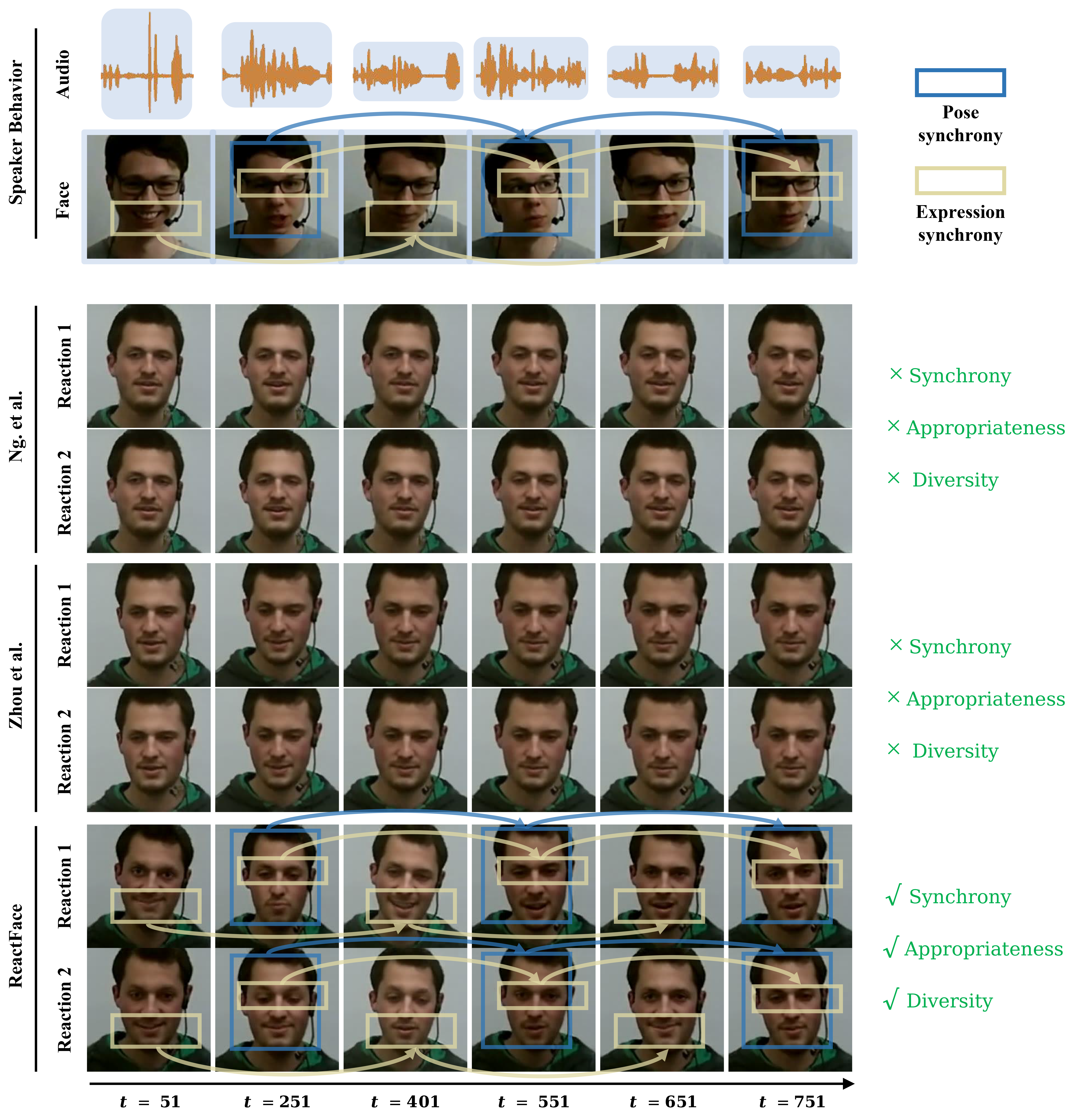} 
            \caption{Comparison between our ReactFace and facial reaction generation methods proposed by Ng. et al \cite{ng2022learning} and Zhou et al. \cite{zhou2022responsive}. The uppermost 
            two rows 
            delineate the auditory, and visual behaviours of a speaker's behavioural clip. The subsequent two rows display a pair of facial reaction sequences generated by \cite{ng2022learning}, in response to the speaker's input. The fifth and sixth rows show a pair of facial reaction sequences generated by \cite{zhou2022responsive}. Finally, the last two rows present diverse, appropriate and synchronised facial reactions generated by our ReactFace.}
    \label{fig:intro}
\end{figure}

\IEEEPARstart{H}{uman} facial reactions play a crucial role in conveying significant non-verbal cues that aid in expressing intentions during human-human interactions \cite{mavridis2015review, van2005identifying}. In dyadic speaker-listener interactions, the facial reaction of a listener to a speaker's behaviour is depending on not solely the stimulus presented by the speaker, but also the contextual factors such as the conversational environment and the listener's individual disposition (\emph{e.g.,} personality \cite{song2022learning,shao2021personality}) \cite{zhai2020sor,mehrabian1974approach,pandita2021psychological}. As a result, facial behaviours expressed for responding to a specific speaker's behaviour might be varied across different listeners and even within the same listener across diverse contextual settings. In other words, \textit{multiple different facial reactions of listeners may be appropriate
in response to a specific stimulus, and thus facial reaction generation can be treated as a 'one-to-many mapping' problem} \cite{song2023multiple}. To this end, it is much more challenging compared with the 'one-to-one' mapping or fitting human behaviour generation problems (with deterministic results), such as speech-driven 3D lip animation \cite{fan2022faceformer}, next sentence prediction \cite{devlin2018bert} 
and deterministic behaviour prediction \cite{zhou2022responsive, pavllo:quaternet:2018}.

Only a limited number of previous studies have investigated the online facial reaction generation task.
Online facial reaction generation is the task of 
producing a listener's facial reaction sequence iteratively during an ongoing conversation with a speaker, in real-time. This approach is fundamentally different from offline generation, where the entire sequence of the speaker's video is processed at once, and the corresponding listener's video is output as a complete sequence. The online method is designed to generate reactions promptly, either autoregressively or segment-by-segment, making it suitable for real-time applications where immediate response is essential.

Early studies \cite{huang2017dyadgan, huang2018generating, huang2018generative} commonly employed conditional generative adversarial networks (CGANs) \cite{mirza2014conditional, choi2018stargan}, which leverage a facial expression image or static facial attributes (\emph{e.g.,} Facial Action Units (AUs)) of the speaker, as conditioning cues. These approaches aimed to reproduce each static facial reaction frame of the corresponding listener. Nevertheless, these approaches overlooked the temporal correlation spanning consecutive frames of the speaker's facial dynamics. Besides, they omitted other speaker behaviour modalities such as audio and text, which are crucial in triggering the listener's facial reactions. Recent facial reaction generation approaches address these limitations by incorporating temporal networks  (\emph{e.g.,} Long-short-term-memory (LSTM) architecture \cite{song2022learning,shao2021personality,zhou2022responsive}), or coupling with supplemental cues extracted from the speaker's auditory \cite{song2022learning,shao2021personality} or textual \cite{ng2022learning} signals. However, these models remain remiss in capturing the \emph{synchrony} between the speaker's behaviour and the corresponding listener's facial response (\textbf{Problem 1}) – a crucial element for creating authentic facial expressions during face-to-face interactions.

Moreover, almost all previous facial reaction generation approaches \cite{huang2018generating, huang2018generative,song2022learning,shao2021personality,nojavanasghari2018interactive,ng2022learning} shared a problematic training strategy, \emph{i.e.}, their training are achieved by directly pairing the input speaker behaviour with the corresponding listener's real facial reaction (called GT facial reaction in this paper). Since facial reaction generation is a 'one-to-many mapping' task, such 'one-to-one mapping' training strategies would cause an ill-posed machine learning problem where similar speaker inputs could paired with different listener facial reaction labels within a given dataset. This would theoretically undermine the feasibility of building reliable models (\textbf{Problem 2}) \cite{song2023multiple}. Consequently, most of these frameworks \cite{huang2018generating, huang2018generative,song2022learning,shao2021personality,nojavanasghari2018interactive} are constrained to generating a single  facial reaction in response to each speaker behaviour. Although some recent models \cite{ng2022learning,jonell2020let} have partially overcame this limitation by generating different facial reactions from a single speaker behaviour input, these approaches failed to consider the \emph{appropriateness} of the generated facial reactions in a given conversational context (\textbf{Problem 3}).

In this paper, we propose \textbf{ReactFace}, an novel online multi-modal framework for generating multiple different yet appropriate, realistic, and temporally synchronized facial reactions in response to audio-visual speaker behaviours. ReactFace pioneers a solution to the MAFRG problem discussed above, with the introduction of two novel strategies. 
Firstly, the \textbf{speaker-listener behaviour synchronization (BS)} strategy synchronises the generated facial reactions with both the speaker speech and visual behaviours across the temporal dimension. This iterative process aligns the facial reactions with the speaker behaviours at each time stamp, ensuring that the generated facial reactions are continuous as well as synchronised with the corresponding speaker behaviours (\textbf{Addressing Problem 1}). 
Additionally, a \textbf{multiple-appropriate facial reaction generation (AFRG)} mechanism is proposed to tackle the 'one-to-many mapping' training problem by re-formulating it as a 'one-to-one mapping' problem. This is achieved by training the model to map a input single speaker behaviour to an distribution representing all appropriate facial reactions (\textbf{Addressing Problem 2}). Consequently, a range of different yet appropriate facial reactions can be sampled based on the obtained distribution learned from the input speaker behaviour (\textbf{Addressing Problem 3}). In summary, the main contributions and novelties of this paper are listed as follows:

\begin{itemize}

     \item We propose the first approach for online multiple appropriate facial reaction generation (online MAFRG), to facilitate the generation of diverse, appropriate, realistic, and temporally aligned facial reactions in response to each speaker behaviour. The main advantage of this online approach lies in its capacity to craft individual facial reaction frames while taking into account the historical and present speaker behaviours. To the best of our knowledge, our ReactFace is the first and only online approach for generating multiple appropriate facial reactions.

    \item We propose a novel speaker-listener behaviour synchronization (BS) module to temporally synchronize the generated facial reactions with the speaker speech and visual behaviours. To the best of our knowledge, this is the first multi-modal synchronization module specifically proposed for the task of facial reaction generation.

    \item The experimental results demonstrate that ReactFace, as posited in this study, outperforms existing state-of-the-art methods in terms of achieving superior interaction synchronization, enhanced diversity, and heightened realism.

\end{itemize}

\begin{table*}[ht]
    \vspace{0.1in}
    \begin{center}
    \resizebox{1.6\columnwidth}{!}{
    \begin{tabular}{ll}
    \hline
         \textbf{Notations}& \textbf{Descriptions} \\
         \hline
        $\bm{\mathrm{B}}^s_{1:t}$ & Speaker audio-facial behaviour \\
        $\bm{\mathcal{X}}^s_{1:t}$  & Raw speaker audio signal \\
        $\bm{\mathrm{F}}^s_{1:t} = \{ \bm{\mathrm{f}}_1^s, \ldots, \bm{\mathrm{f}}_t^s \}$  &  2D speaker face image sequence \\
        $\bm{\mathrm{R}}^s_{1:t} = \{ \bm{\mathrm{r}}_1^s, \ldots, \bm{\mathrm{r}}_t^s \}$  & 3D speaker face image sequence \\ \hline

        $\bm{A}^s_{1:kt} = \{ \bm{a}_1^s, \ldots, \bm{a}_{kt}^s \}$  & Extracted speaker speech feature sequence \\ 
        $\bm{\tilde{F}}^s_{1:t} = \{ \bm{\tilde{f}}_1^s, \ldots, \bm{\tilde{f}}_t^s\}$ & Extracted speaker facial feature sequence \\
        $\bm{\bar{F}}^s_{1:t} = \{ \bm{\bar{f}}_1^s, \ldots, \bm{\bar{f}}_t^s\}$  & Aligned speaker facial feature sequence\\  
       $\hat{\bm{\mathrm{R}}}^s_{1:t} = \{ \bm{\hat{\mathrm{r}}}_1^s, \ldots, \bm{\hat{\mathrm{r}}}_t^s\}$ & 3D speaker face image sequence prediction \\ \hline \hline

        $\bm{\mathrm{F}}^l_{1:t} = \{ \bm{\mathrm{f}}_1^l, \ldots, \bm{\mathrm{f}}_t^l\} $  &  2D facial reaction image sequence \\
    
        $\bm{\mathrm{R}}^l_{1:t} = \{ \bm{\mathrm{r}}_1^l, \ldots, \bm{\mathrm{r}}_t^l \}$  & 3D facial reaction image sequence \\ \hline

        $\bar{\bm{F}}^l_{1:t-w} \{ \bar{\bm{f}}_{1}^l, \ldots, \bar{\bm{f}}_{t-w}^l\}$ &  Aligned listener facial feature sequence \\

        $\bm{R}_{t-w+1:t}^l = \{ \bm{r}_{t-w+1}^l, \ldots, \bm{r}_t^l\}$ & Raw 3D facial reaction sequence prediction  \\

        $\bm{\bar{R}}^l_{t-w+1:t} = \{ \bm{\bar{r}}_{t-w+1}^l, \ldots, \bm{\bar{r}}_t^l\}$ & Visually synchronised 3D facial reaction sequence prediction \\

        $\hat{\bm{\mathrm{R}}}^l_{t-w+1:t} = \{ \hat{\bm{\mathrm{r}}}_{t-w+1}^l, \ldots, \hat{\bm{\mathrm{r}}}_t^l\}$ & Finally synchronised 3D facial reaction sequence prediction \\
        
        $\bm{P}_{t-w+1:t}^l = \{ \bm{p}_{t-w+1}^l, \ldots, \bm{p}_{t}^l\}$ & Positional encodings\\  
        
        $\bm{Z}_{t-w+1:t} = \{ \bm{z}_{t-w+1}, \ldots, \bm{z}_{t}\}$   & Sampled latent vectors representing an appropriate facial reaction \\ \hline

    \hline
    \end{tabular}}
    \end{center}
    \caption{Notations used in this paper.}
    \label{tab:notation}
    \vspace{-0.2in}
\end{table*}

\section{Related Work}
\label{sec:related_work}

\noindent 
This section begins with a recap of previous approaches to facial reaction generation and facial behavior generation in Sec. \ref{subsec:facial reaction genration}, followed by a review of the literature on other non-verbal behaviors, such as gesture and body motion generation, in Sec. \ref{subsec:non-verbal reaction genration}. Next, it discusses conditional generative models in Sec.~\ref{subsec:Conditional Generation} and examines the modality alignment techniques utilized in Sec.~\ref{subsec:monditional generation}.

\subsection{Automatic Facial Reaction Generation}
\label{subsec:facial reaction genration}

\textbf{Facial Reaction Generation.} Existing approaches attempt to reproduce the listener's GT facial reactions from the input speaker behaviour. For example, early studies frequently \cite{huang2017dyadgan,huang2018generating, huang2018generative,nojavanasghari2018interactive} extended the conditional GAN \cite{mirza2014conditional,goodfellow2020generative} to predict each listener's GT facial reaction under a one-on-one virtual interview scenario, which takes each speaker's facial image as the input and aim to reproduce a frame-level GT facial reaction (sketches) expressed by the corresponding listener. Recent facial reaction studies \cite{ng2022learning,zhou2022responsive} frequently represent facial motions using 3DMM (3D Morphable Model) coefficients (\emph{e.g.}, expression and pose coefficients) to better visualize the facial muscle movements. Meanwhile, speaker's speech and audio behaviours were also additionally employed as the input to deliver richer verbal and non-verbal speaker behavioural cues \cite{ng2022learning,song2022learning,shao2021personality}, which are crucial in triggering listeners' facial reactions. For example, Ng et al. \cite{ng2022learning} proposed a VQ-VAE-based framework that considers the nondeterministic nature of human facial reactions by learning a set of codebook vectors. Song et al. \cite{song2022learning,shao2021personality} search for a network whose architecture and weights are specifically adapted to each target listener's person-specific facial reactions in response to the speaker's audio and facial non-verbal behaviour. As discussed in \cite{song2023multiple,mehrabian1974approach}, human facial reaction has an uncertain nature, \emph{i.e.}, the same/similar behaviours expressed by speakers could trigger different facial reactions across not only various subjects but also the same subject under different contexts. Thus, it is problematic to train models that aims to reproduce the corresponding listener's GT facial reaction only from each speaker behaviour sequence. As a result, a publicly Grand Challenge \cite{song2023react2023} focusing on Multiple Appropriate Facial Reaction Generation (MAFRG) has been organised. It is important to note that since the submission of our paper, numerous MAFRG approaches have been developed \cite{song2024react, xu2023reversible, nguyen2024vector, dam2024finite, zhu2024perfrdiff, nguyen2024multiple, liu2024one, liang2023unifarn, yu2023leveraging, hoque2023beamer}. However, it is not fair to directly compare these methods with ours, as some of them \cite{xu2023reversible,dam2024finite, nguyen2024multiple,nguyen2024vector} are offline MAFRG approaches, while others \cite{nguyen2024multiple, hoque2023beamer} are built upon our ReacFace framework.

\textbf{Facial Animation.} The proposed facial reaction generation approach also involves facial animation, which has attracted considerable attention in recent years \cite{cao2016real,fried2019text, le2012live, kim2018deep,lahiri2021lipsync3d,li2013realtime}. In particular, speech-driven face generation models aim to automatically animate vivid facial expressions of a 2D portrait \cite{zhou2021pose,kim2018deep,li2013realtime, wen2020photorealistic} or 3D avatar face sequence \cite{fan2022faceformer,lahiri2021lipsync3d} from a human speech signal. For example, StyleGAN-V \cite{skorokhodov2022stylegan} and StyleFaceV \cite{qiu2022stylefacev} are designed to generate a temporally coherent facial motion sequence with the same identity. Moreover, facial expression \cite{wang2020mead,liang2022expressive,ji2021audio} or head pose \cite{zhou2021pose,liang2022expressive} representations are additionally employed as the condition signal in some approaches, in order to manually control the generated face sequence. However, given a subject's speech, all these works can only generate a face sequence of the corresponding subject rather than the conversational partner's facial reactions.

\subsection{Non-verbal Human body/gesture Behaviour Generation}
\label{subsec:non-verbal reaction genration}

Besides the facial behaviour generation, a large number of studies also have been devoted to the domain of the neural body and gesture motion generation models \cite{tevet2022human, petrovich2021action, TEACH:3DV:2022, pavllo:quaternet:2018, habibie2017recurrent, barsoum2018hp, pavllo:quaternet:2018, kanazawa2019learning} over the preceding decades. They frequently represent the generated human motions in the form of 3D skeletons \cite{yuan2020dlow,lee2019dancing, lin2018human}, video frames \cite{walker2017pose}, or 3D parameters  (\emph{e.g.,} SMPL \cite{tevet2022human,petrovich2022temos,TEACH:3DV:2022}). This evolution encompasses a spectrum of generative models such as Generative Adversarial Networks (GANs) \cite{barsoum2018hp,lin2018human}, variational autoencoder (VAE) \cite{petrovich2022temos}, and normalizing flows \cite{yuan2020dlow} as well as the recent entrant of diffusion models \cite{tevet2022human}.

These motion synthesis approaches can be divided into two primary categories: unconstrained generation and conditioned generation. Unconstrained approaches \cite{yan2019convolutional, zhang2020perpetual, zhao2020bayesian} are usually achieved by networks that model a latent space of all possible motions. For example, Yan et al. \cite{yan2019convolutional} proposed a convolutional sequence generation framework that can produce long actions transformed from a sequence of latent vectors sampled from a Gaussian process. In contrast, conditioned networks are defined to yield different outputs depending on conditional signals, including music \cite{lee2019dancing, aristidou2022rhythm}, speech \cite{bhattachar, ginosar2019learning}, action \cite{guo2020action2motion}, and text \cite{lin2018human, petrovich2022temos, guo2022generating}. However, the samples generated by these deterministic generative models \cite{pavllo:quaternet:2018, aksan2019structured} and non-deterministic generative models \cite{yan2019convolutional, zhang2020perpetual, petrovich2022temos} frequently suffer from insufficient diversity problem. In this paper, we consider generation diversity in interaction scenarios, where we offer a fresh idea by transforming from one-to-one supervision to one-to-many supervision, based on which the well-trained model achieved high diversity in its outputs (illustrated in Fig.~\ref{fig:AFRRL}).

\subsection{Conditional Generative Models}
\label{subsec:Conditional Generation}

Recent techniques for conditional generation have been developed to incorporate various modalities, ensuring that synthesized results align with the instructions provided by these modality signals \cite{sohn2015learning}. Early approaches, such as conditional variational autoencoders (CVAEs) \cite{sohn2015learning} and conditional generative adversarial networks \cite{sohn2015learning}, leverage class labels to distinguish image domains and guide generated image samples to possess domain-specific properties. For example, CycleGAN was proposed to translate images from a source domain to a target domain, while the StarGAN series \cite{choi2018stargan, choi2020stargan} further succeeded in translating images from a source domain to multiple target domains. 
Some techniques embed discrete class labels as conditional signals in the generation process, either by directly concatenating class labels with the input \cite{choi2018stargan, sohn2015learning} or by using conditional normalization techniques \cite{choi2020stargan, karras2019style}. Building on these successful practices, conditional diffusion models incorporate class information (\emph{e.g.,} class labels or text) into normalization layers and guide the generation process using classifier gradients. Further research has shown that guidance can be derived from the generative model itself without a classifier, a method known as classifier-free guidance. This advancement unlocks various forms of conditional signals, leading to conditional generative models \cite{zhang2023adding, rombach2022high} that control synthesized samples with multiple modalities, such as text, edge maps, human pose skeletons, segmentation maps, depth, and normals.

\begin{figure*}[t!]
    \centering
    \includegraphics[width=2.0\columnwidth]{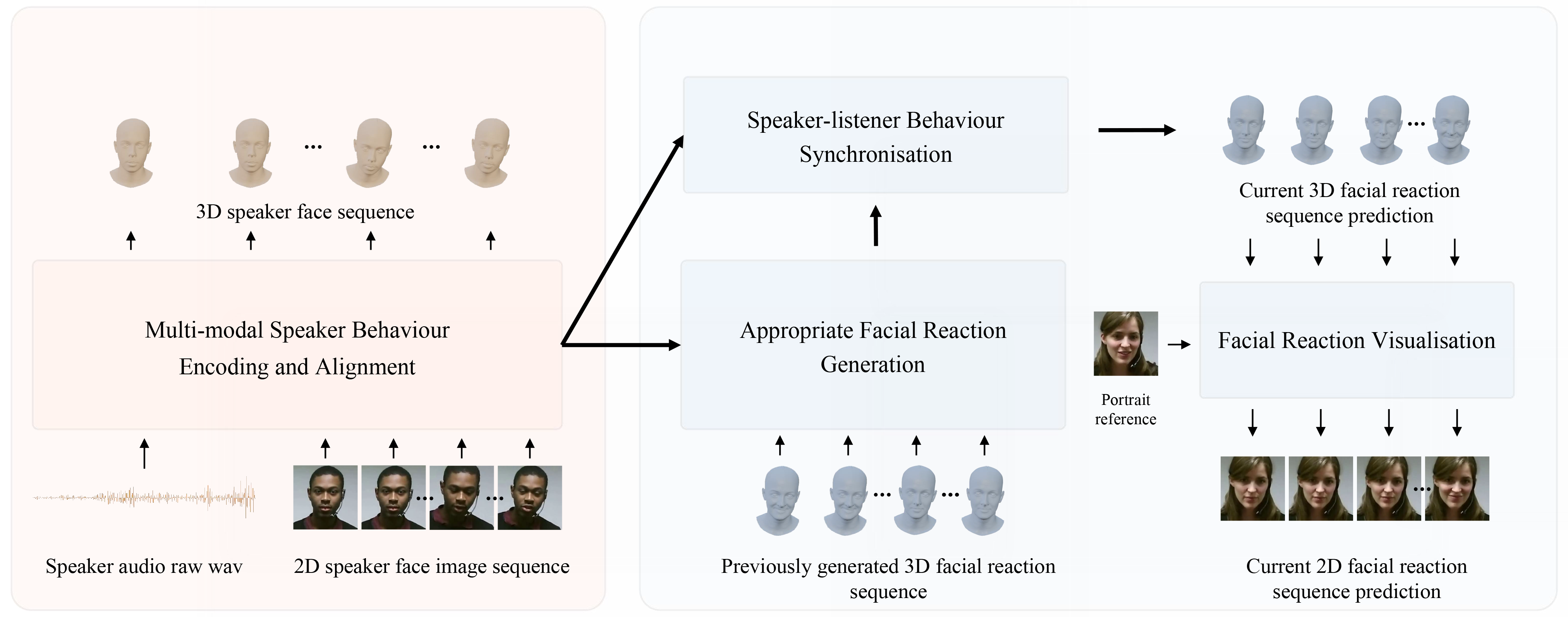} 
        \caption{The pipeline of the proposed ReactFace model.}
    \label{fig:pipline}
\end{figure*}

\subsection{Modality Alignment in Generative Models}
\label{subsec:monditional generation}

Modality alignment is crucial for generative models processing diverse data types like text, images, audio, and video. This alignment ensures coherent content generation and manipulation across modalities, but faces challenges due to semantic and dimensional differences. 
To address the semantic disparity, researchers have turned to contrastive learning techniques. Works such as Zhang et al. \cite{zhang2023adding} and Kim et al. \cite{kim2022diffusionclip} have built upon the foundational approach introduced by Radford et al. \cite{radford2021learning} in CLIP. This method utilizes positive and negative pairs to learn aligned representations, proving particularly effective in text-to-vision synthesis tasks.
The dimensionality mismatch between modalities presents another significant hurdle. Recent advancements, exemplified by Zhang et al. \cite{zhang2023adding} and Rombach et al. \cite{rombach2022high}, have leveraged cross-modal attention mechanisms. These allow models to focus on relevant features within each modality, effectively reducing dimensionality while preserving crucial information. 
A straightforward yet effective approach to unifying different modalities has been the adoption of multimodal transformers, as demonstrated by Tsai et al. \cite{tsai2019multimodal}. This method, which represents different modalities as unified tokens, has found wide application across various generative tasks. It has been successfully employed in text-to-image generation \cite{zhang2023adding}, text-to-motion synthesis \cite{tevet2022human, TEACH:3DV:2022}, and even multi-modal generation \cite{zhan2024anygpt}, showcasing its versatility and effectiveness in bridging the gap between diverse data types.
For temporal alignment, innovative attention variants have emerged. Press et al. \cite{press2021train} introduced Attention with Linear Biases (ALiBi), while Fan et al. \cite{fan2022faceformer} proposed Biased Cross-Modal Attention with periodic position encoding. These approaches incorporate temporal alignment biases into the attention mechanism, ensuring meaningful matches between query and key modalities in the temporal dimension.
However, these temporal alignment methods fail to model the temporal interaction between two conversation participants. For instance, the point-to-point audio-visual alignment in Fan et al. \cite{fan2022faceformer} does not account for audio-visual influences from previous frames. In this paper, we propose a novel approach to modality interaction over time, which can effectively model the multi-modal signal exchange between two subjects engaged in an ongoing dyadic interaction.

\section{Problem Formulation}
\label{subsec:method_task_defin}

At the time $t$, the goal of the \textbf{online MAFRG task} is to develop a machine learning (ML) model $\mathcal{M}$ that can generate multiple, different, but appropriate spatio-temporal facial reactions in response to the given speaker behaviour $\bm{\mathrm{B}}^s_{1:t}$, which can be formulated as:
\begin{equation}
\label{eq:task_defin}
  \hat{\bm{\mathrm{F}}}^l_{t-w+1:t} = \mathcal{M}(\bm{\hat{\mathrm{F}}}^l_{1:t-w}, \bm{\mathrm{B}}^s_{1:t})
\end{equation}
where $\hat{\bm{\mathrm{F}}}^l_{1:t-w}$ denotes the previously predicted/generated 2D facial reaction sequence, and $w$ represents the facial reaction delay caused by the execution of the human cognitive processes \cite{card1986model}. Importantly, the well-trained model $\mathcal{M}$ is expected to be able to generate multiple different but appropriate facial reaction sequences from each input speaker behaviour. For more details, please refer to \cite{song2023multiple}.

\section{Methodology}
\label{sec:method}

\noindent This section introduces our online MAFRG model ($\text{ReactFace}$), which aims to generate multiple, different, yet appropriate, realistic, and synchronized facial reactions from each input spatio-temporal speaker behavior. To account for the time delay caused by human cognitive processes in facial reaction generation, we follow \cite{song2022learning,shao2021personality,ng2022learning} in defining a small time window $w$ (with $w=8$ used in this paper). Subsequently, the speaker's audio-visual behavior $\bm{\mathrm{B}}^s_{1:t}$ can be split into two parts as:
\begin{equation}
\bm{\mathrm{B}}^s_{1:t} = \{ \bm{\mathrm{B}}_{1:t-w}^s,  \bm{\mathrm{B}}_{t-w+1:t}^s \},
\end{equation}
where $\bm{\mathrm{B}}_{1:t-w}^s = \{\bm{\mathcal{X}}^s_{1:t-w}, \bm{\mathrm{F}}^s_{1:t-w} \}$ denotes the previous audio-facial speaker behavior expressed over the time period $[1, t-w]$, and $\bm{\mathrm{B}}_{t-w+1:t}^s = \{\bm{\mathcal{X}}^s_{t-w+1: t}, \bm{\mathrm{F}}^s_{t-w+1:t} \}$ denotes the current audio-facial speaker behavior expressed over the time period $[t-w+1, t]$. Thus, $\bm{\mathrm{B}}^s_{1:t}$ can also be denoted as:
\begin{equation}
    \bm{\mathrm{B}}^s_{1:t} = \{ \bm{\mathcal{X}}^s_{1:t}, \bm{\mathrm{F}}^s_{1:t} \},
\end{equation}
where $\bm{\mathcal{X}}$ and $\bm{\mathrm{F}}$ represent the original audio signal and 2D facial image sequence, respectively. Then, our ReactFace model generates a 3D facial reaction segment ($\hat{\bm{\mathrm{R}}}_{t-w+1:t}^l$) consisting of $w$ frames (corresponding to the time period $[t-w+1, t]$) based on: (1) the previously generated/predicted facial reaction sequence $\hat{\bm{\mathrm{R}}}_{1:t-w}^l$; (2) the previous speaker behavior $\bm{\mathrm{B}}_{1:t-w}^s$; and (3) the current speaker behavior $\bm{\mathrm{B}}_{t-w+1:t}^s$, which can be formulated as:
\begin{equation}
    \hat{\bm{\mathrm{R}}}_{t-w+1:t}^l = \mathcal{M}
    (\hat{\bm{\mathrm{R}}}_{1:t-w}^l, \bm{\mathrm{B}}_{1:t-w}^s, \bm{\mathrm{B}}_{t-w+1:t}^s).
\end{equation}

In the following subsections, we first present the entire framework of the ReactFace in Sec. \ref{subsec:pipeline}, providing a high-level overview for the pipeline. Next, we provide a detailed explanation of the main technical contributions of the ReactFace model: the multiple facial reaction generation strategy (Sec.~\ref{subsec:Multiple}) and the speaker-listener behaviour synchronisation module (Sec.~\ref{subsec:synchronisation}).

\subsection{The ReactFace Framework}
\label{subsec:pipeline}

The entire framework of the proposed ReactFace is illustrated in Fig.~\ref{fig:pipline}, which is made up of four main modules: (i) a \textbf{Multi-modal speaker behaviour encoding and alignment (MSBEA) module} that encodes the input speaker audio-facial behaviour as a set of aligned and semantic meaningful speaker behaviour embeddings; (ii) an \textbf{Appropriate facial reaction generation (AFRG) module}, which predicts a distribution of appropriate facial reactions based on the speaker behaviour embeddings, allowing for the sampling of multiple appropriate facial reaction embeddings; 
(iii) a \textbf{Speaker-listener behaviour synchronisation (SLBS) module} that ensures the generated facial reaction embeddings to be synchronised with the corresponding speaker behaviour; and (iv) a \textbf{Facial reaction visualisation (FRV) module} that further decodes a set of 3D and 2D facial frames representing the corresponding sampled appropriate facial reaction. The full pipeline is described as follows:

\begin{figure*}[t!]
    \centering
    \includegraphics[width=2.0\columnwidth]{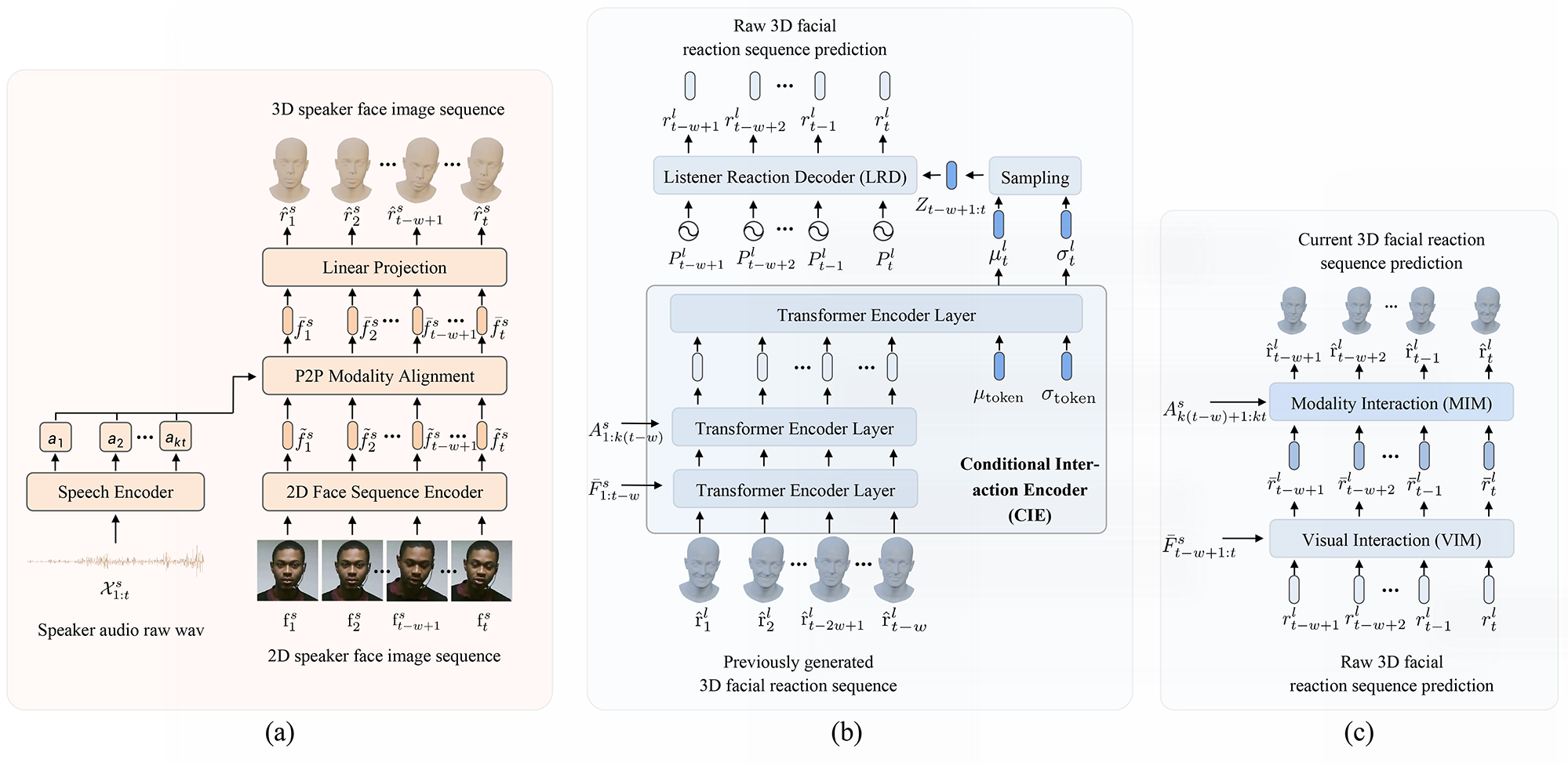} 
        \caption{Illustration of: (a) multi-modal speaker behaviour encoding and alignment  (MSBEA) module; (b) appropriate facial reaction generation (AFRG) module; and (c) speaker-listener behaviour synchronisation (SLBS) module.}
    \label{fig:model_details}
\end{figure*}

\begin{figure}[t]
    \centering
    \includegraphics[width=1\columnwidth]{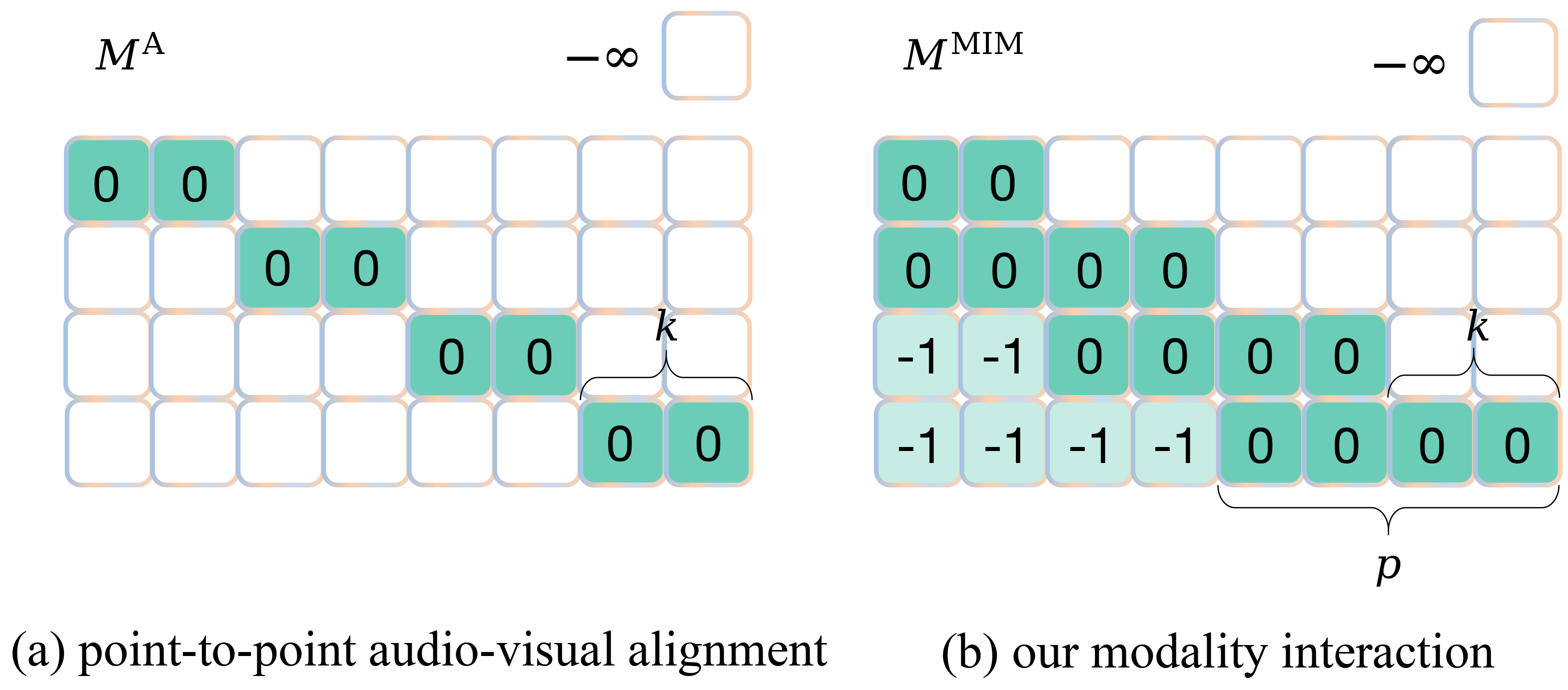} 
        \caption{Comparison of: (a) point-to-point (P2P) audio-visual alignment proposed in Faceformer \cite{fan2022faceformer} and (b) our proposed modality interaction. P2P audio-visual alignment  primarily establishes corrections solely between a given visual frame and its corresponding audio frames at the identical timestamp (achieved through the assignment of zero attention weights) while effectively precluding the visual frame from attending to any other frames (achieved through the assignment of negative infinity attention weights). Our proposed modality interaction mechanism establishes both long-range and short-range relationships between a visual frame at a specific timestamp and audio frames occurring at both the current and previous time instances. This approach significantly enriches the capacity to capture intermodal dependencies and contextual nuances across disparate frames within the audio-visual data stream.}
    \label{fig:point-to-point}
\end{figure}

\begin{itemize}

    \item \textbf{(i) Multi-modal speaker behaviour encoding and Alignment (MSBEA):} Given a speaker behaviour $\bm{\mathrm{B}}_{1:t}^s$, the MSBEA module first encodes it as a set of verbal speech and non-verbal facial behaviour embeddings. To achieve this, a speech encoder A-Enc (\emph{i.e.}, a state-of-the-art pre-trained speech model wav2vec2.0~\cite{baevski2020wav2vec}) is leveraged to encode the raw audio signal $\bm{\mathcal{X}}^s_{1:t}$ as a set of speaker speech embeddings $\bm{A}^s_{1:kt}$, where $k$ denotes that sampling rate for obtaining speech frames is $k$ times over the sampling rate for obtaining facial frames, which depends on the audio encoder's setting. Meanwhile, a 2D facial sequence encoder F-Enc (\emph{i.e.,} a network consisting of 3D convolutional layers and transformer encoder layers) encodes all speaker facial frames $\bm{\mathrm{F}}^s_{1:t}$ as a set of speaker facial embeddings $\bm{\tilde{F}}^s_{1:t}$. This can be formulated as:
    \begin{equation}
    \begin{split}
        \bm{A}^s_{1:kt} &= \text{A-Enc}(\bm{\mathcal{X}}^s_{1:t}) \\
        \bm{\tilde{F}}^s_{1:t} &= \text{F-Enc}(\bm{\mathrm{F}}^s_{1:t} ) 
    \end{split}
    \end{equation}
    Then, a modality alignment module (\emph{i.e.,} a cross-attention transformer with an specific alignment bias explained in \cite{fan2022faceformer}) is proposed to align the speaker facial embeddings with the corresponding speech embeddings in a point-to-point manner (please refer to Fig.~\ref{fig:point-to-point} for the explanation of this term), resulting in a set of aligned facial embeddings $\bm{\bar{F}}^s_{1:t}$ as: 
    \begin{equation}
        \bm{\bar{F}}^s_{1:t} =  \text{PMA}(\bm{A}^s_{1:kt}, \bm{\tilde{F}}^s_{1:t})
    \end{equation}
    where $\text{PMA}$ denotes the 'Point-to-Point Modality Alignment', \emph{i.e.,} the operation that aligns facial feature at each time stamp only with the corresponding speech features. To ensure the aligned speaker facial embeddings being semantic meaningful, the MSBEA module also enforces the $\bm{\bar{F}}^s_{1:t}$ to contain information that allows the corresponding 3D human-style facial reaction sequence $\bm{\mathrm{\hat{R}}}_{1:t}^s$ can be reconstructed from it. Please refer to Fig.~\ref{fig:model_details} (a) for the visual illustration of the MSBEA module.

    \item \textbf{(ii) Appropriate facial reaction generation (AFRG):} The AFRG involves applying a Conditional Interaction Encoder (CIE) to predict a distribution (\emph{i.e.,} Gaussian distribution $\mathcal{N}(\bm{\mu}^l_{t}, \bm{\sigma}^l_{t})$) representing multiple appropriate facial reaction segments (\emph{i.e.,} each segment contains $w$ frames) at the current time $[t-w+1,t]$ based on: (1) aligned previous speaker speech and facial behaviour embeddings ($\bm{A}^s_{1:k(t-w)}$ and $\bm{\bar{F}}^s_{1:t-w}$); (2) previously predicted and aligned facial reaction embeddings $\bm{\mathrm{\hat{R}}}^l_{1:t-w}$; and (3) two learnable tokens (\emph{i.e.,} $\bm{\mu}_\text{token}$ and $\bm{\sigma}_\text{token}$). This can be formulated as:
    \begin{equation}
    \begin{split}
        \mathcal{N}(\bm{\mu}^l_{t}, \bm{\sigma}^l_{t}) =  \text{CIE}(& \bm{A}^s_{1:k(t-w)}, \bm{\bar{F}}^s_{1:t-w}, \\
       & \bm{\mathrm{\hat{R}}}^l_{1:t-w}, \bm{\mu}_\text{token}, \bm{\sigma}_\text{token}) 
    \end{split}
    \end{equation}    
    Consequently, an embedding  $\bm{z}_{t}  \sim \mathcal{N}(\bm{\mu}^l_{t}, \bm{\sigma}^l_{t})$ can be sampled from the learned distribution and we then use a temporal linear interpolation to get a set of embeddings
    $\bm{Z}_{t-w+1:t} = \{\bm{z}_{t-w+1}, \ldots, \bm{z}_{t} \}$ representing an appropriate facial reaction segment at the time $[t-w+1,t]$, where each component $\bm{z}_{\tau} \in \bm{Z}_{t-w+1:t}$ denotes an appropriate facial reaction frame of the time stamp $\tau$. \textbf{Importantly, this process allows multiple different embeddings that represent various appropriate facial reactions to be sampled, while the use of previously predicted facial reaction embeddings enables the generated current facial reaction to be continuous with previously generated facial reactions.} Then, a listener reaction decoder (LRD) decodes the sampled $\bm{Z}_{t-w+1:t}$ as $w$ 3D facial reaction frames, which can be formulated as:
    \begin{equation}
        \bm{{R}}_{t-w+1:t}^l = \text{LRD}(\bm{Z}_{t-w+1:t}, \bm{P}_{t-w+1:t}^l) 
    \end{equation} 
    where $\bm{P}_{t-w+1:t}^l =  \{ \bm{P}_{t-w+1}^l, \ldots,  \bm{P}_{t}^l\}$ are positional encodings representing temporal positions in the whole generated clip and $\bm{R}_{t-w+1:t}^l =  \{ \bm{r}_{t-w+1}^l, \ldots,  \bm{r}_{t}^l\}$ denote the $w$ sets of 3D facial reaction features representing $w$ current 3D facial reaction frames. These generated 3D frames can be further used to render 2D listener identity-aware face image sequence $\bm{\hat{\mathrm{F}}}^l_{1:t}$. Please refer to Fig.~\ref{fig:model_details} (b) for the visual illustration of the AFRG module.

    \item \textbf{(iii) Speaker-listener behaviour synchronisation (SLBS).} the SLBS module synchronises the generated facial reactions for every $w$ frames. At the time $t$, it synchronises the generated $w$ current 3D facial reaction frames $\bm{R}^l_{t-w+1:t}$ with the corresponding aligned current speaker facial embeddings $\bm{\bar{F}}^s_{t-w+1:t}$ and speech embeddings $\bm{A}^s_{k(t-w)+1:kt}$, which can be formulated as:
    \begin{equation}
        \bm{\hat{\mathrm{R}}}^l_{t-w+1:t} = \text{SLBS}(\bm{R}^l_{t-w+1:t}, \bm{\bar{F}}^s_{t-w+1:t}, \bm{A}^s_{k(t-w)+1:kt})
    \end{equation}
    where $\bm{\hat{\mathrm{R}}}^l_{t-w+1:t} = \{\bm{\hat{\mathrm{r}}}_{t-w+1}^l, \ldots, \bm{\hat{\mathrm{r}}}_{t}^l \}$ denotes the synchronised current 3D facial reaction ($w$ frames). Please check Fig.~\ref{fig:model_details} (c) for the visual illustration of the SLBS module.

    \item \textbf{(iv) Facial reaction visualisation.} To translate from generated 3D facial reactions (represented by 3DMM coefficients) to 2D facial frames, we re-train the PIRender (\cite{ren2021pirenderer}) to make it adapt to our used FaceVerse \cite{wang2022faceverse} 3DMM. With the produced 3DMM coefficients and a reference portrait of a specific facial identity, this render is able to output corresponding 2D frames of a human listener facial reaction clip.
    
\end{itemize}

\subsection{Appropriate Facial Reactions Generation}
\label{subsec:Multiple}

As described above, the AFRG module consists of three main blocks: (i) a Conditional Interaction Encoder (CIE) that predicts an appropriate facial reaction distribution $\mathcal{N}(\bm{\mu}^l_{t}, \bm{\sigma}^l_{t})$; (ii) a sampling block that outputs a latent embedding $\bm{z}_{t} \sim \mathcal{N}(\bm{\mu}^l_{t}, \bm{\sigma}^l_{t})$ and gets a set of latent samples  $\bm{Z}_{t-w+1:t} = \{\bm{z}_{t-w+1}, \ldots, \bm{z}_{t} \} $ describing an appropriate facial reaction from the predicted distribution; and (iii) a listener facial reaction decoder (LRD) that decodes the sampled facial reaction embeddings as $w$ raw 3D facial reaction frames $\bm{R}^l_{t-w+1:t} = \{\bm{r}_{t-w+1}^l, \ldots, \bm{r}_{t}^l \}$.

The \textbf{CIE} is a variational encoder that is composed of three transformer encoder layers. As illustrated in Fig.~\ref{fig:model_details} (b), it takes: (i) $\bm{\mathrm{\hat{R}}}^l_{1:t-w}$, which represents previously generated facial reaction features that are further synchronized with previous speech ($\bm{A}^s_{1:k(t-w)}$) and facial ($\bm{\bar{F}}^s_{1:t-w}$) features of the corresponding speaker behaviour, and (ii) a pair of learnable tokens ($\bm{\mu}_{token}$ and $\bm{\sigma}_{token}$) representing reaction distribution prior, as the input. Then, it outputs a pair of vectors $\bm{\mu}^l_{t}$ and $\bm{\sigma}^l_{t}$, which are respectively the mean and standard deviation of a Gaussian distribution $\mathcal{N}(\bm{\mu}^l_{t}, \bm{\sigma}^l_{t})$ describing multiple appropriate facial reactions in response to the input speaker behaviour. In this process, $\bm{\bar{F}}^l_{1:t-w}$, $\bm{\mu}_{token}$ and $\bm{\sigma}_{token}$ are jointly input as transformer tokens to the CIE. Consequently, three types of tokens can be produced, where two of them ($\bm{\mu}^l_{t}$ and $\bm{\sigma}^l_{t}$) are employed for the sampling. Note that the positions of $\bm{\mu}^l_{t}$ and $\bm{\sigma}^l_{t}$ are the same as their prior $\bm{\mu}_{token}$ and $\bm{\sigma}_{token}$, where an attention mask is utilized to only calculate attention scores between distribution tokens ($\bm{\mu}^l_{t}$ and $\bm{\sigma}^l_{t}$) and the corresponding facial reaction features $\bm{\bar{F}}^l_{1:t-w}$, respectively. This makes the CIE generating each facial reaction by considering the temporal relationship between it and its previous facial reactions (\emph{i.e.,} previously generated facial reactions).

Once the distribution $\mathcal{N}(\bm{\mu}^l_{t}, \bm{\sigma}^l_{t})$ is obtained, the \textbf{sampling block} first samples a latent embedding $\bm{z}^{*}_{t}$ that represents the $t_\text{th}$ frame of the predicted facial reaction. In particular, the embedding $\bm{z}^{*}_{t}$ is further updated as $\bm{z}_t$ based on an embedding $\bm{z}_{t-w}$ that represents a previously predicted facial reaction frame (\emph{i.e.,} ${(t-w)}_\text{th}$ frame). This can be defined as:
\begin{equation}
    \bm{z}_t = \alpha \cdot \bm{z}_{t-w} + (1 - \alpha) \cdot \bm{z}^{*}_t
\end{equation}
where $\alpha$ is a momentum parameter to avoid the generated facial reaction frames to be dramatically changed from adjacent previously predicted facial reaction frames. Then, linear interpolation is employed to produce $w$ smooth and continuous latent vectors $\bm{Z}_{t-w+1: t} = \{\bm{z}_{t-w+1}, \ldots, \bm{z}_{t} \}$ corresponding to the facial reaction predictions at the time $[t-w+1,t]$, which can be formulated as: 
\begin{equation}
    \bm{z}_\tau = \bm{z}_{t-w} + (\tau  - (t-w))(\bm{z}_{t} -  \bm{z}_{t-w} )/w, \tau \in (t-w, t]
\end{equation}
This process further allows the smooth transition for the $w$ generated facial reaction frame from the facial display of the $({t-w})_\text{th}$ frame to that of the $t_\text{th}$ frame. In short, the $w$ facial reaction frames are all generated based on the previously predicted $({t-w})_\text{th}$ facial reaction frame, which ensures the facial reactions generated for the time $[t-w+1,t]$ to be smooth and continuous with the previously generated facial reactions.

Finally, the \textbf{LRD} is introduced to generate $w$ 3D facial reaction frames from the sampled embeddings $\bm{Z}_{t-w+1:t} = \{\bm{z}_{t-w+1}, \ldots, \bm{z}_{t} \} $. It performs cross-attention between a sequence of positional encodings $\bm{P}_{t-w+1: t}^l$ (\emph{i.e.,}  treated as the keys and values) and the $\bm{Z}_{t-w+1: t}$ (\emph{i.e.,}  treated as the queries), which is computed as:
\begin{equation}
\begin{aligned}
\bm{R}^l_{t-w+1:t}=\operatorname{softmax}\!\left(\!\!\frac{\bm{Z}_{t-w+1: t}(\bm{P}_{t-w+1: t}^l)^{T}}{\sqrt{d_p}}\right)\!\bm{P}_{t-w+1: t}^l
\end{aligned}
\end{equation}
where $\sqrt{d_p}$ denotes the channel number of the key $\bm{P}_{t-w+1: t}$. These positional encodings are obtained based on the sinusoidal functions proposed in \cite{vaswani2017attention} as:
\begin{equation}
\begin{split}
& \bm{P^{l}}_{(t,2i)} =  \text{sin}(t/10000^{2i/d_p})\\  
& \bm{P^{l}}_{(t,2i+1)} =  \text{cos}(t/10000^{2i/d_p})
\end{split}
\label{eq:pe}
\end{equation}
where each positional encoding $\bm{P}_t^l$ embeds a specific temporal cue into its corresponding frame token, which makes the generation of each facial reaction frame considering its global position at the time $(0, t]$. This way, each facial reaction embedding in $\bm{R}^l_{t-w+1:t}$ would be aware of its temporal position within the entire facial reaction sequence.

\subsection{Speaker-listener Behaviour Synchronisation}
\label{subsec:synchronisation}

Since human facial reaction is a spatio-temporal signal, which is highly correlated with the corresponding speaker behaviour in the temporal dimension, properly synchronising the generated facial reactions with the corresponding speaker behaviours is a key to ensure them to be appropriate and realistic.

As shown in Fig.~\ref{fig:model_details} (c), the \textbf{SLBS} module is made up of two models: a \textbf{Visual Interaction Model (VIM)} and a \textbf{Modality Interaction Model (MIM)}. Specifically, the VIM first synchronises the generated current raw 3D facial reactions (\emph{i.e.,} the embeddings $\bm{R}^l_{t-w+1:t}$) with the corresponding aligned speaker facial embeddings $\bm{\bar{F}}^s_{t-w+1:t}$ produced by the MSBEA module as:
\begin{equation}
    \bm{\bar{R}}^l_{t-w+1:t}  =  \text{VIM}( \bm{R}^l_{t-w+1:t}, \bm{\bar{F}}^s_{t-w+1:t})
\end{equation}
Then, the MIM is introduced to further synchronise these visually synchronised facial reactions $\bm{\bar{R}}^l_{t-w+1:t}$ with the corresponding speaker speech embeddings (\emph{i.e.,} speaker speech embeddings $\bm{A}^s_{k(t-w)+1:kt}$), respectively. This process can be formulated as:
\begin{equation}
\begin{split}
    &\bm{\hat{\mathrm{R}}}^l_{t-w+1:t} = \text{MIM}( \bm{\bar{R}}^l_{t-w+1:t},\bm{A}^s_{k(t-w)+1:kt})
\end{split}
\end{equation}
This way, the final generated facial reaction frames $\bm{\hat{\mathrm{R}}}^l_{t-w+1:t}$ would be synchronised with both non-verbal facial behaviours and verbal speech behaviours expressed by the corresponding speaker.

\begin{figure}[t!]
    \centering
    \includegraphics[width=1\columnwidth]{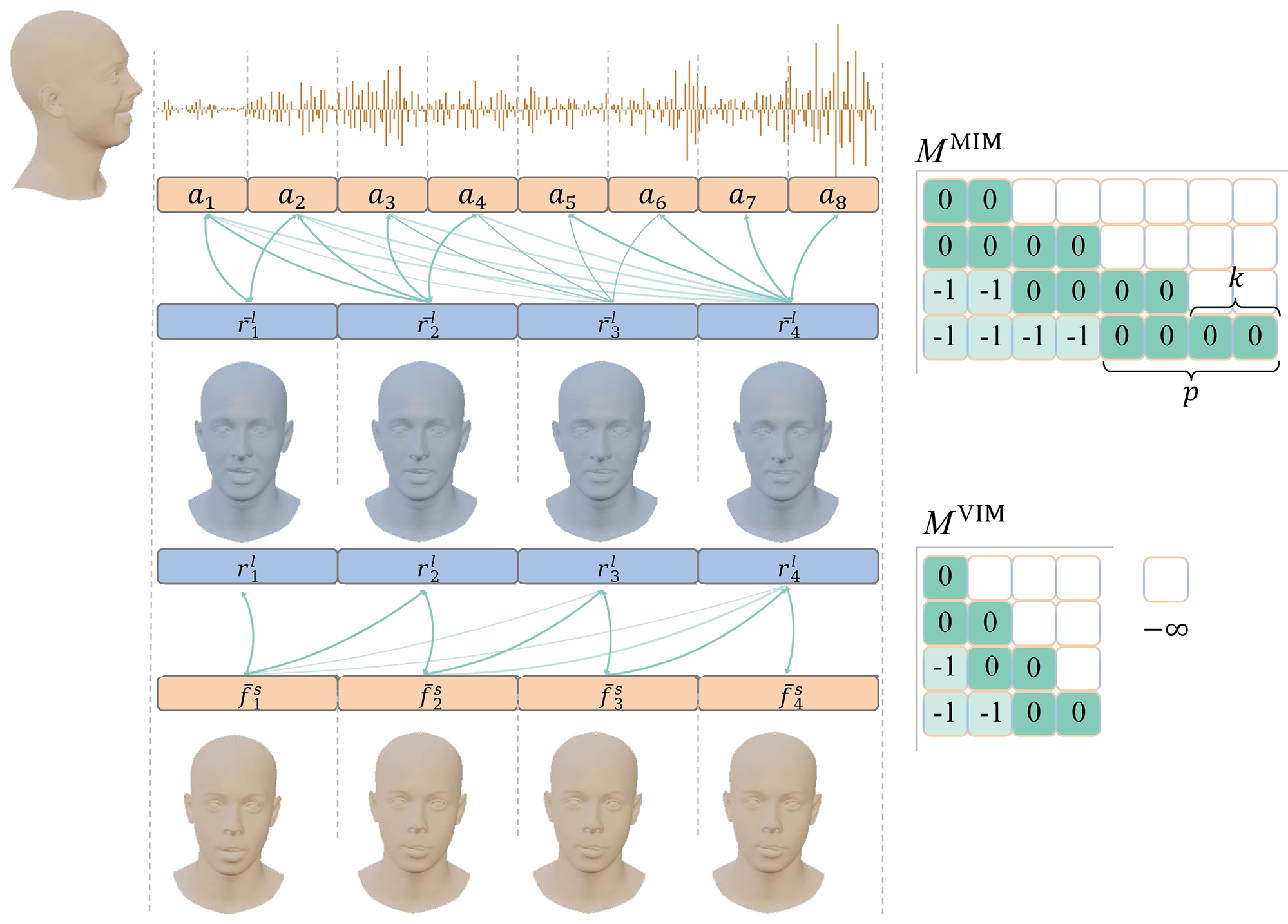} 
        \caption{Illustration of visual interaction and modality interaction}
    \label{fig:multimodal-align}
\end{figure}

In this paper, both VIM and MIM are cross-attention operations, where a novel alignment bias $\bm{M}^b$ is proposed under the assumption that the speaker behaviours that are temporally closer to the time $i$ would have more influence on the facial reaction $\bm{R}^l_{i}$ generated at the time $i$. In particular, this alignment bias is proposed for two purposes: (i) synchronising the generated facial reactions with the corresponding speaker behaviours (both non-verbal and verbal behaviours) in the temporal dimension; and (ii) ensuring the synchronisation of each facial reaction frame $\bm{r}_{i}^l$/$\bm{\bar{r}}_{i}^l$ is only influenced by historical speaker behaviours as well as the previously generated facial reactions, \emph{i.e.,} frames previous to the position $i$. This novel cross-attention operation is formulated as:
\begin{equation}
\label{eq:mutimodal}
\begin{aligned}
\operatorname{Attention}(\bm{Q},\!\bm{K}\!,\!\bm{V}\!,\!\bm{M}^b)=\operatorname{softmax}\!\left(\!\!\frac{\bm{Q}\bm{(K)}^{T}}{\sqrt{d_k}}\!\!+\!\!\bm{M}^b\!\!\right)\!\bm{V}
\end{aligned}
\end{equation}
where the speech/facial behaviours embedding expressed by the speaker are input as the keys $\bm{K}$ and values $\bm{V}$ in VIM/MIM, while the generated facial reactions are treated as the queries $\bm{Q}$.

The $\bm{M}^b$ is a matrix that has similar forms in VIM and MIM, which can be denoted as $\bm{M^{\text{VIM}}}$ and $\bm{M^{\text{MIM}}}$. Here, the component in $i_\text{th}$ row and $j_\text{th}$ column (\emph{i.e.,} $\bm{M}^{\text{VIM}}_{i, j} $ and $\bm{M}^{\text{MIM}}_{i, j} $) of $\bm{M^{\text{VIM}}}$ and $\bm{M^{\text{MIM}}}$ can be computed as:
\begin{equation}
\label{eq:b^vim} 
\begin{split}
& \bm{M}^{\text{VIM}}_{i, j} \!= \!\left\{\begin{array}{ll} \!\!\!\!-\left\lfloor\frac{i-j}{p}\right\rfloor, \!& \!\!  \!j \leq i, \\ \! \!\!\!-\infty,\! &\! \! \!   j>i\end{array}\right. \! \\
& \bm{M}^{\text{MIM}}_{i, j}\!=\!\left\{\begin{array}{ll}
\!\!\!\!-\left\lfloor\frac{ki-j}{kp}\right\rfloor, \!& \!  \! \!  j \leq k \times i \\
\! \!\!\!-\infty, \!&\!  \! \!j > k \times i
\end{array}\right.
\end{split}
\end{equation}
where $i$ and $j$ denote that the attention is conducted between the $i_\text{th}$ facial reaction frame and the $j_\text{th}$ speaker facial frame/speech frame, \emph{i.e.,} $k$ represents that the sampling rate for obtaining speech frames is $k$ times over the sampling rate for obtaining facial frames. Here, $p$ is a parameter that divides all processed speaker facial/speech frames and the corresponding facial reaction frames into several units, where each unit contains $p$ facial frames and $k \times p$ consecutive speech frames. 
This is because nearby frames (within a window) usually share similar behavior patterns. Setting parameter $p$ considers this human behaviour nature and prevents excessively small attention weights to distant previous context, ensuring  sufficient  attention to long-range information.
Subsequently, the attention operation treats all frames in each unit equally. Fig.~\ref{fig:multimodal-align} further illustrates our temporal bias matrix, where the upper triangle of $\bm{M}^b$ contains negative infinity to ensure that each facial reaction frame is synchronised only based on previous and current speaker behaviours, without considering future information, which is not known in real-world online applications. More importantly, Eqa. \ref{eq:b^vim} demonstrated that in both VIM and MIM, the synchronisation of the $i_\text{th}$ facial reaction frame only takes the speaker speech and facial behaviour frames that prior or equal to it into consideration. This is because if $j>i$ / $j>ki$, $M^{\text{VIM}}_{i, j} = -\infty$ / $M^{\text{MIM}}_{i, j} = -\infty$, and subsequently the output of the attention operation defined in Eqa. \ref{eq:mutimodal} would be zero. In summary, the proposed parameter $\bm{M}^b$ determines the influence of speaker speech and facial frames on the synchronization of the $i_\text{th}$ facial reaction frame, based on their temporal proximity. Specifically, the speaker speech and facial frames that exhibit closer temporal distances to the $i_\text{th}$ facial reaction frame would have more substantial impacts on its synchronization.

Different from previous work \cite{fan2022faceformer} that only individually conducts point-to-point (p2p) audio-visual alignment at each time stamp, aiming at matching human's verbal signals to its visual lip movements synchronously in talking head generation task, the proposed SLBS module aligns each generated facial reaction frame with both current and past speaker speech/facial behaviours. This allows our SLBS module to capture long-range dependencies between speaker audio-visual behaviour and the generated facial reactions, as listener's facial reactions are always influenced by both current and past interactions in dyadic interaction.

\subsection{Loss functions and training strategy}
\label{method:training}

To allow the proposed framework to be easily reproduced, a simple end-to-end training strategy is proposed in Sec. \ref{subsec:training}, where several loss functions are jointly employed to ensure the well-trained model to generate appropriate, diverse, realistic and synchronised facial reactions (Sec. \ref{subsec:loss functions}).

\subsubsection{Training Strategy}
\label{subsec:training}

\begin{algorithm}[t]
    \caption{Training Strategy}
    \begin{algorithmic}
    \label{al:training_strategy}
        \REQUIRE{2D speaker facial image sequence $\bm{\mathrm{F}}_{1:T}^s$, 3D speaker facial image sequence $\bm{\mathrm{R}}_{1:T}^s$, Raw speaker audio signal $\bm{\mathcal{X}}^s_{1:t}$, the number of sampling $M$,  window size $w$}
        \STATE $NT$ $\gets \lceil T/w \rceil$
        \FOR{$i=1$ to $NT$}
        \STATE $t \gets i \times w$ ;
        \STATE Obtain $\bm{\hat{\mathrm{R}}}_{1:t}^s$, $\bm{A}_{1:t}^s$, and $\bm{\bar{F}}_{1:t}^s$ though MSBEA module;
        \STATE Obtain $\bm{R}_{t-w+1:t}^l$ though AFRG module;
        \STATE Synchronize $\bm{R}_{t-w+1:t}^l$ with $\bm{\bar{F}}_{t-w+1:t}^s$ and $\bm{A}_{k(t-w+1):kt}^s$ through SLBS module and obtain $\bm{\hat{\mathrm{R}}}^l_{t-w+1:t}$;
        \ENDFOR
        \STATE Calculate speaker face reconstruction loss as in Eq.~\eqref{eq: rec_loss_speaker}.

        \STATE Calculate appropriate facial reaction reconstruction loss, KL divergence loss and temporal smooth loss as in Eq.~\eqref{eq: rec_loss_stg2}, Eq.~\eqref{eq: kl_loss} and Eq.~\eqref{eq: smooth_loss}, respectively.
        
        \FOR{$j=2$ to $M$}
            \STATE Sample new reaction predictions  $\bm{\hat{\mathrm{R}}}^l_{j, 1:T}$ 
        \ENDFOR
        \STATE Calculate diversity loss as in Eq.~\eqref{eq:diversity_loss}
    \end{algorithmic}
\end{algorithm}

We propose an end-to-end strategy to train our ReactFace model, which is achieved by jointly optimising five loss functions as:
\begin{equation}
\label{eq: total_loss}
\mathcal{L} =  \mathcal{L}_\text{rec}^a + \mathcal{L}_\text{rec}^s + \lambda_\text{kl} \mathcal{L}_\text{kl} + \lambda_\text{smo} \mathcal{L}_\text{smo} +  \lambda_\text{div} \mathcal{L}_\text{div} 
\end{equation}
where $\mathcal{L}_\text{rec}^a$ represents an \textit{appropriate facial reaction reconstruction loss} that focuses on enforcing the model to generate appropriate and realistic (semantic meaningful) facial reactions; 
$\mathcal{L}_\text{rec}^s$ is a reconstruction loss that supervises the 3D speaker facial reaction sequence generation achieved by our MSBEA module; $\mathcal{L}_\text{kl}$ is a \textit{Kullback-Leibler (KL) divergence loss} that constrains the learned appropriate facial reaction distribution; $\mathcal{L}_\text{smo}$ is a \textit{temporal smooth loss} that avoids jitters in the generated facial frames; and $\mathcal{L}_\text{div}$ denotes the \textit{energy-based diversity loss} that enforces the model to generate diverse facial reactions. Meanwhile, a set of weights $\lambda_\text{kl}, \lambda_\text{smo}, \lambda_\text{div}$ are also employed to decide the relative importance of the aforementioned five loss functions, respectively, where $\lambda_\text{kl}$ and $\lambda_\text{div}$ are used to balance the trade-off between realism/appropriateness and diversity of the generated facial reactions. For clarity, we present the pseudo-code in Algorithm~\ref{al:training_strategy} to outline our training strategy.

\subsubsection{Loss Functions}
\label{subsec:loss functions}

To ensure that MSBEA module encodes human interpretable cues from the input speaker facial behaviour, we first employ a reconstruction loss to supervise its outputs $\bm{\hat{\mathrm{R}}}_{1:t}^s$, which is defined as:
\begin{equation}
\label{eq: rec_loss_speaker}
 \mathcal{L}_\text{rec}^{s} = \mathscr{L}_1(\bm{\hat{\mathrm{R}}}_{1:T}^s, \bm{\mathrm{R}}_{1:T}^s) 
\end{equation}
where $\mathscr{L}_1$ denotes the smooth L1 loss \cite{girshick2015fast} and  $T$ denotes the total number of frames in the sequence.

\begin{figure}[t!]
    \centering
    \includegraphics[width=1.0\columnwidth]{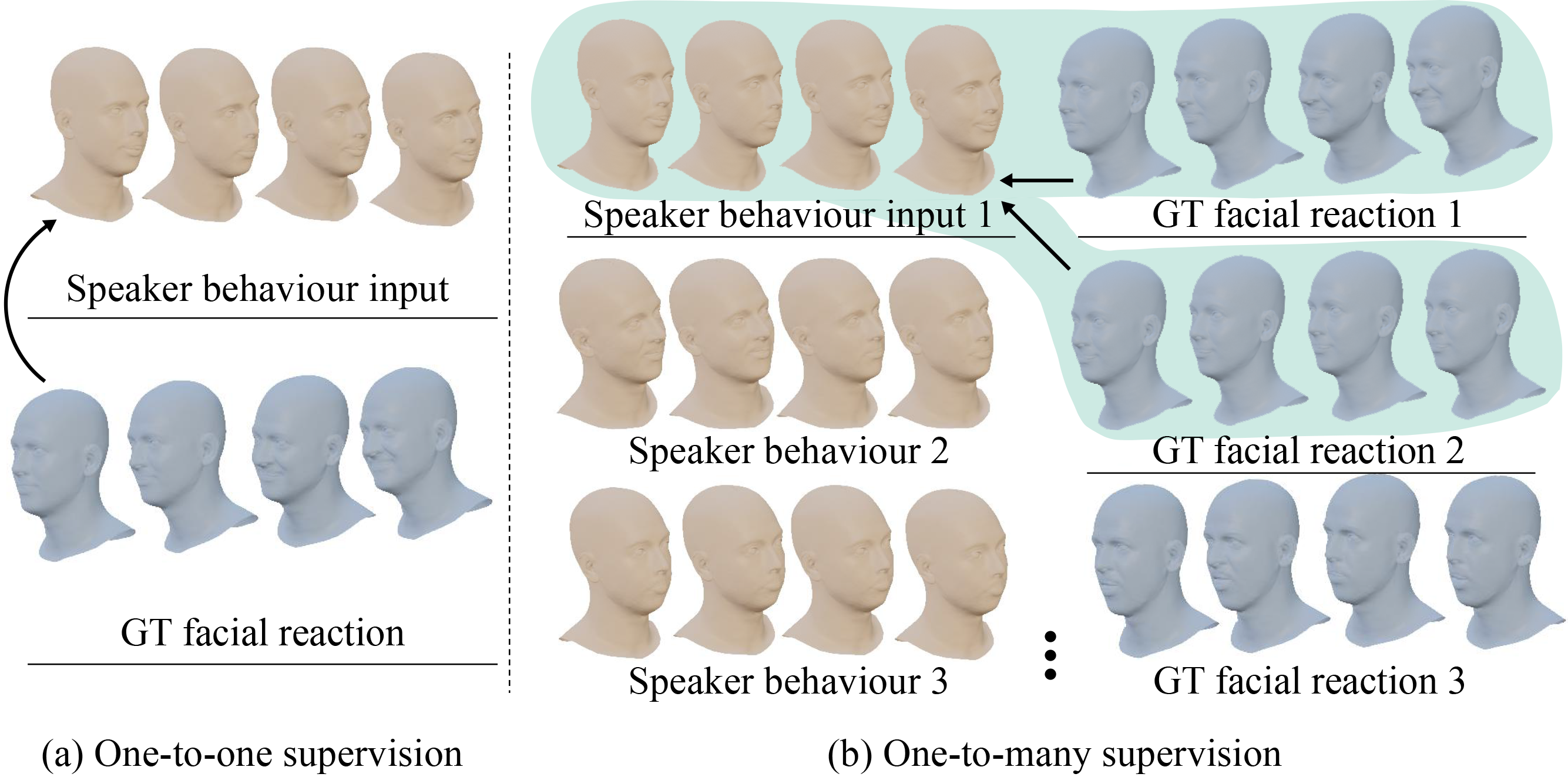} 
        \caption{Illustration of appropriate facial reaction reconstruction loss. We replace (a) one-to-one GT reaction supervision with (b)  supervision by appropriate reactions (GT 1, GT 2, and so on). } 
    \label{fig:AFRRL}
\end{figure}

To enforce the model to learn a distribution $\mathcal{N}(\mu^l_{t}, \sigma^l_{t})$ that can represent multiple appropriate and realistic facial reactions in response to an input speaker behaviour, the \textbf{appropriate facial reaction reconstruction loss ($\mathcal{L}_\text{rec}^a$)} is computed between the generated 3D facial reaction sequence (\emph{i.e.,} $\bm{\hat{\mathrm{R}}}^l_{1:T}$) and the 3D facial sequence corresponding to its most similar appropriate real facial reaction $\bm{\mathrm{R}}^{l}_{1:T}(i)$ in the training set as:
\begin{equation}
\label{eq: rec_loss_stg2}
\mathcal{L}_\text{rec}^{a} = \min(\{ \mathscr{L}_1 (\bm{\hat{\mathrm{R}}}^{l}_{1:T},\bm{\mathrm{R}}^{l}_{1:T}(i))|  \bm{\mathrm{R}}^{l}_{1:T}(i) \in \mathbb{N}(\bm{\mathrm{B}}^{s}_{1:T})\} )
\end{equation}
where $\bm{\mathrm{R}}^{l}_{1:T}(i)$ denotes an appropriate real facial reaction in response to the input speaker behaviour, which has the highest similarity to the generated facial reaction among all appropriate real facial reactions of the input speaker behaviour; $\bm{\mathrm{B}}^{s}_{1:T}$ is the input speaker behaviour and $\mathbb{N}(\bm{\mathrm{B}}^{s}_{1:T})$ represents a set of real facial reactions that are appropriate for responding to $\bm{\mathrm{B}}^{s}_{1:T}$. Here, the appropriate real facial reactions for each speaker is obtained based on the automatic appropriate labelling strategy proposed in \cite{song2023multiple}. While each speaker behaviour may lead to multiple different facial reactions, the proposed appropriate facial reaction reconstruction loss $\mathcal{L}_\text{rec}^a$ does not propagate a large loss value if the generated facial reaction is similar to any of the appropriate real facial reactions, \emph{i.e.,} even the generated facial reaction is not similar to the corresponding listener's GT facial reaction. Please refer to Fig. \ref{fig:AFRRL} and \cite{song2023multiple} for more detailed definition of appropriate facial reactions. 

Since one goal of the developed model is to generate multiple different facial reactions from each input speaker behaviour, we employ an \textbf{energy-based diversity loss ($\mathcal{L}_\text{div}$)} \cite{yuan2020dlow} to compute the difference between each pair of generated facial reactions (\emph{i.e.,} we sample $M$ facial reactions for each speaker behaviour) as:
\begin{equation}
\label{eq:diversity_loss}
\mathcal{L}_\text{div} = \frac{1}{M(M-1)} \sum_{i=1}^{M}\sum_{j \neq i}^{M} \text{exp}(-\frac{\|\bm{\hat{\mathrm{R}}}^{l}_{i, 1:T} - \bm{\hat{\mathrm{R}}}^{l}_{j, 1:T}
 \|_2^2}{\sigma_d})
\end{equation}
where an RBF kernel with scale $\sigma_d$ is used for measuring the distance between a pair of facial reactions ($\bm{\hat{\mathrm{R}}}^{l}_{T,i}$ and $\bm{\hat{\mathrm{R}}}^{l}_{T,j}$), and $M=3$ at the training stage (we follow \cite{song2023react2023} to set $M=10$ at the testing stage, \emph{i.e.,} generating 10 facial reactions from each speaker behaviour). This way, minimizing the diversity loss $\mathcal{L}_\text{div}$ allows the model to generate facial reactions toward a lower-energy (\textbf{higher diversity}). In summary, the use of the above two loss functions achieves two goals: (1) making the learned distribution $\mathcal{N}(\bm{\mu}^l_{t}, \bm{\sigma}^l_{t})$ to well represent corresponding appropriate real facial reactions; and (2) enabling model to generate diverse facial reactions from each speaker behaviour.

To constrain distribution parameters $\bm{\mu}^l$ and $\bm{\sigma}^l$ for the purpose of maintaining realism of the generated facial reactions, we apply the canonical \textbf{Kullback-Leibler (KL) divergence loss ($\mathcal{L}_{kl}$)} to make the learned $\mathcal{N}(\bm{\mu}^l_{t}, \bm{\sigma}^l_{t})$ approaching the standard normal distribution $\mathcal{N}(0, I)$ as:
\begin{equation}\label{eq: kl_loss}
\mathcal{L}_\text{kl} =  \text{KL}(\bm{\phi}, \bm{\omega} )
\end{equation}
This strategy has been widely used in \cite{petrovich2021action, guo2022generating, petrovich2022temos, athanasiou2022teach} to regularize the learned Gaussian distribution $\bm{\phi} = \mathcal{N}(\bm{\mu}^l_t, \bm{\sigma}^l_t)$ with the standard normal distribution $\bm{\omega}  = \mathcal{N}(0, I)$.

Finally, to avoid jitters in the generated facial frames, we utilize a \textbf{temporal smooth loss ($\mathcal{L}_\text{smo}$)} to constrain facial motion variations between adjacent frames as:
\begin{equation}
\label{eq: smooth_loss}
\mathcal{L}_\text{smo} = \frac{1}{(T-2)} \sum_{t= 2}^{T} \mathscr{L}_1((\hat{\bm{\mathrm{r}}}_{t} - \hat{\bm{\mathrm{r}}}_{t-1} ) ,(\hat{\bm{\mathrm{r}}}_{t-1} - \hat{\bm{\mathrm{r}}}_{t-2} ) )
\end{equation}
This loss term constrains the change velocities ($\hat{\bm{\mathrm{v}}}_{t} = (\hat{\bm{\mathrm{r}}}_{t} - \hat{\bm{\mathrm{r}}}_{t-1} )$ and $\hat{\bm{\mathrm{v}}}_{t-1} = (\hat{\bm{\mathrm{r}}}_{t-1} - \hat{\bm{\mathrm{r}}}_{t-2} )$) of adjacent frames to be close, making translation between adjacent frames to be more smooth.

\begin{figure*}[t!]
    \centering
    \includegraphics[width=2\columnwidth]{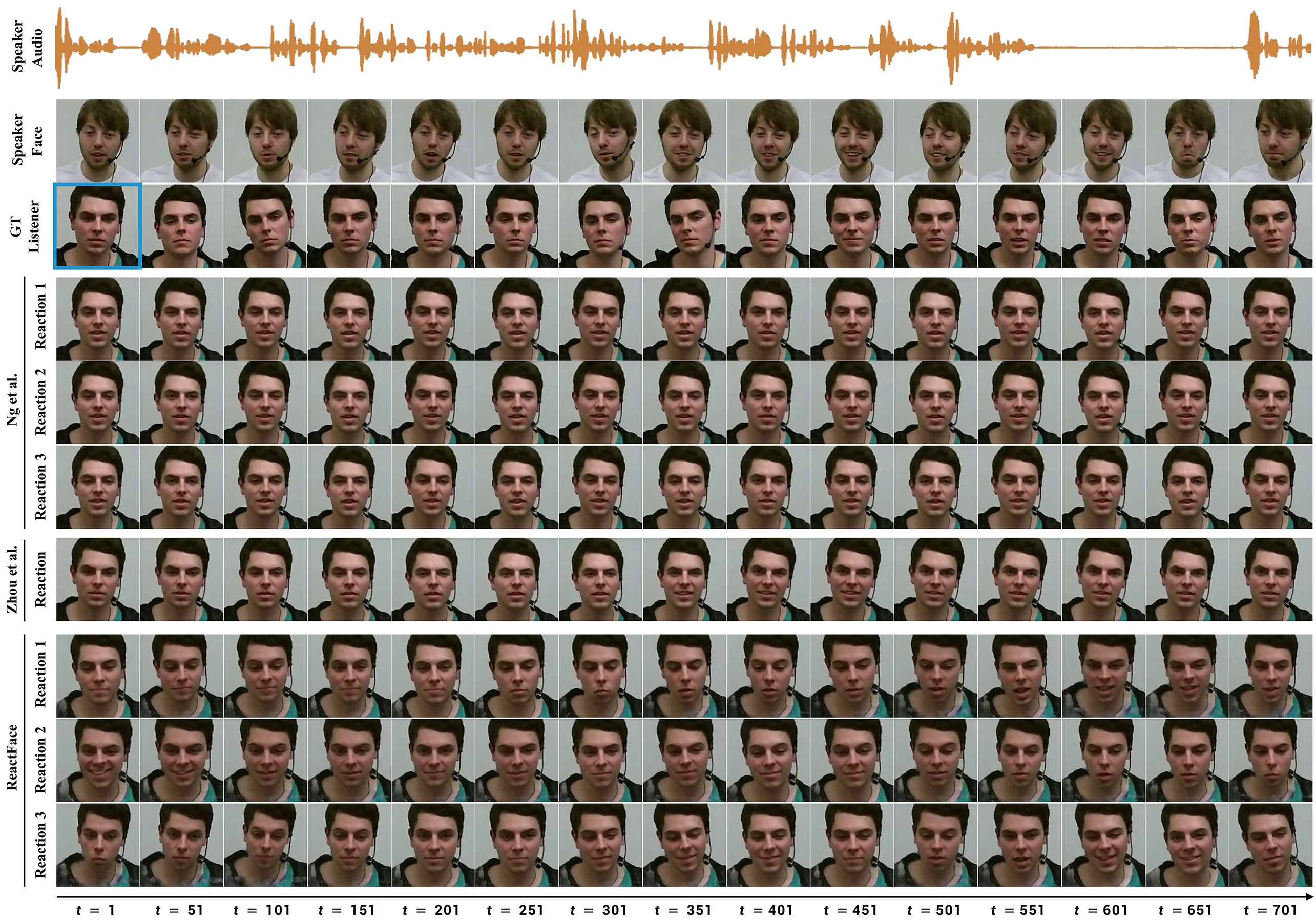} 
            \caption{Qualitative results: Our approach (ReactFace) can generate multiple different but appropriate and synchronised  facial reactions in response to each audio-visual speaker behaviour (including its face frames and audio). The portrait reference (the 1st GT frame) is framed in blue.}
    \label{fig:vis_diff_sample}
\end{figure*}

\begin{figure*}[t!]
    \centering
    \includegraphics[width=2\columnwidth]{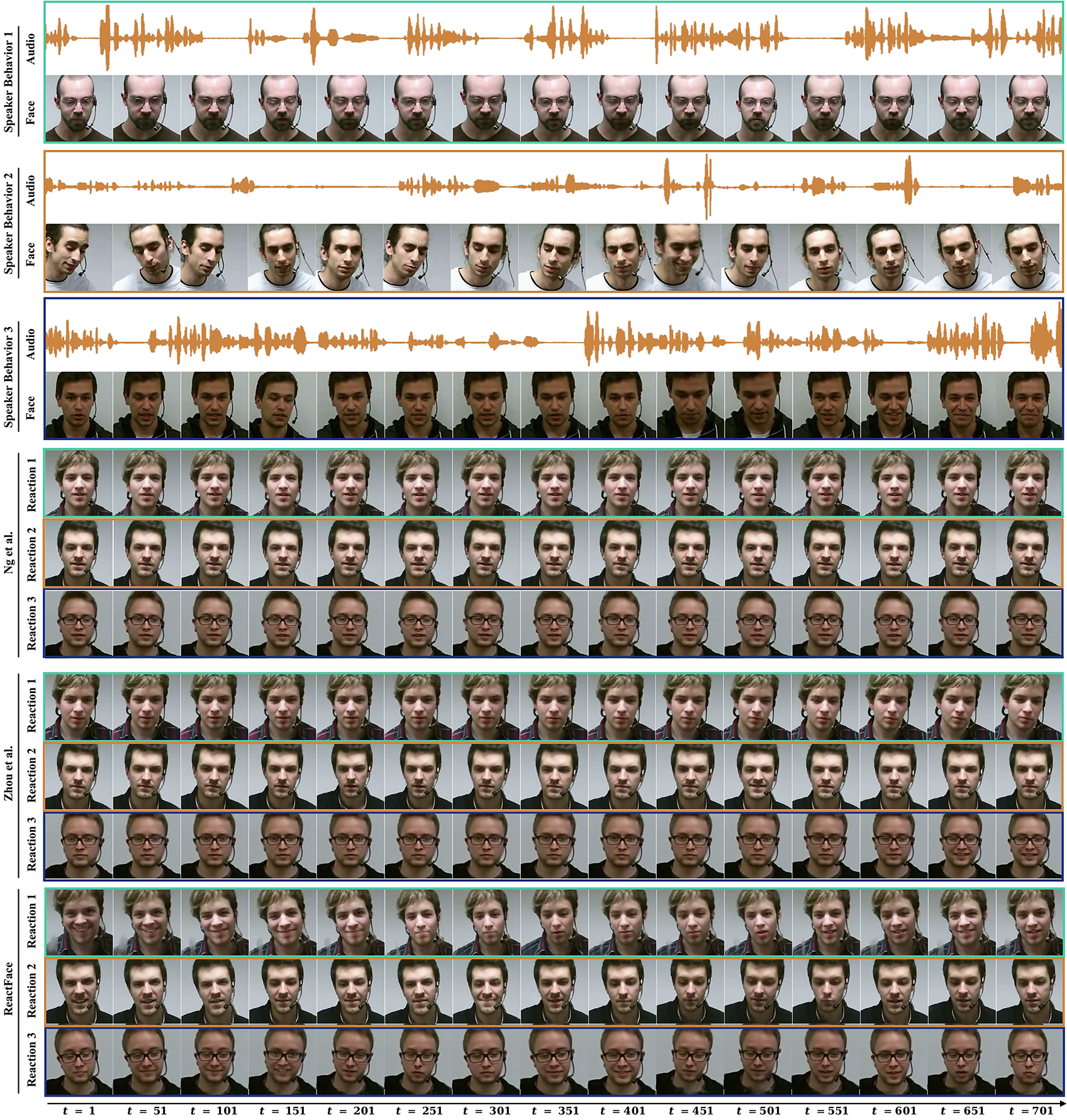} 
            \caption{Qualitative results: Our approach (ReactFace) can generate facial reactions with high diversity in response to different speaker behaviours.}
    \label{fig:vis_diff_speaker}
\end{figure*}

\section{Experiments}
\label{sec:exp}

In this section, we first present the details of the employed datasets in Sec. \ref{subsec:dataset}. All experimental setups are provided in Sec. \ref{subsec:Setup}. We then compare qualitative and quantitative performances between our ReactFace and several re-produced baselines in Sec. \ref{subsec:Qualitative Results} and Sec. \ref{subsec:Quantitative Results}. Finally, ablation and perceptual study results, and failure cases are discussed in Sec.~\ref{subsec:ablation}, Sec.~\ref{subsec:perceptual}, and Sec.~\ref{sec:failure_cases}, respectively.

\subsection{Datasets}
\label{subsec:dataset}

We evaluate our ReactFace on a hybrid video conference dataset provided by REACT2023 challenge \cite{song2023react2023}, which is made up of 2,962 dyadic interaction sessions (1,594 training examples, 562 validation examples and 806 test examples) coming from two video conference datasets: RECOLA \cite{ringeval2013introducing} and NOXI \cite{cafaro2017noxi}. Each session is 30s long and contains a pair of audio-visual clips describing two subjects' interaction behaviours. The objective appropriate facial reaction annotations are also obtained using the strategy proposed in \cite{song2023react2023}. Due to the license issue, this paper does not include UDIVA \cite{palmero2021context} dataset. Please refer to the challenge site \footnote{https://sites.google.com/cam.ac.uk/react2023/home} for more detailed description of the employed dataset, license and appropriateness labels used in our experiments.

\begin{table*}[t]\centering
 \caption{\label{tb:compare_hb} 
 Quantitative results comparison between the ReactFace, baselines and four state-of-the-art method. The best result in each column is marked in bold.} 
 \resizebox{1\textwidth}{!}{
\begin{tabular}{lccccccc}
\toprule 
\multirow{2}{*}{\textbf{Method}} & \multicolumn{3}{c}{\textbf{Diversity}}  & \multicolumn{1}{c}{\textbf{Realism}} & \multicolumn{1}{c}{\textbf{Appropriateness}}  &   \multicolumn{1}{c}{\textbf{Synchrony}} & \multicolumn{1}{c}{\textbf{Speed}}\\
\cmidrule(r){2-4} \cmidrule(r){5-5}  \cmidrule(r){6-6}  \cmidrule(r){7-7} \cmidrule(r){8-8} 
    &  \textbf{FRDvs} ($\uparrow$) &  \textbf{FRVar} ($\uparrow$)  &  \textbf{FRDiv} 
 ($\uparrow$) & \textbf{FRRea} ($\downarrow$)   & \textbf{FRCorr} ($\uparrow$) & \textbf{FRSyn} ($\downarrow$)  & \textbf{FPS} ($\uparrow$)  \\ \midrule

     GT   & 0.0374 & 0.0120 & 
  -   &   282.03
    &  9.480  & 
  48.46 & 
-  \\
Mirror & 0.0374 & 0.0120 & 0 & 282.03 & 0.936 & 42.65 & 
-  \\
Random & 0.0415 & 0.0202 & 0.0041 & 477.49 & 0.127 & 45.82 & 
- \\
NN motion & 0.0420 & 0.0199 & 0 & 452.38 & 0.334 & 46.90 & 
15.36  \\
NN audio & 0.0464 & 0.0218 & 0 & 496.25 & 0.017 & 47.67 & 
3.32 \\
\midrule
Trans-AE & 0.0063 & 0.0003 & 0 & 599.35 & \textbf{0.245} & 45.01 & 
119.73 \\
LFI \cite{jonell2020let} & 0.0247 & 0.0109 & 0.0138 & 734.06 & 0.048 & 45.92 & 
 24.15 \\
Ng et al. \cite{ng2022learning} & 0.0015 & 0.0102 & 0.0003 & 691.24 & 0.059 & 45.70 & 
 91.75 \\
Zhou et al. \cite{zhou2022responsive} & 0.0106 & 0.0039 & 0 & 527.47 & 0.104 & 45.24 & 
 \textbf{150.14}\\
\cellcolor{yellow}ReactFace (Ours) & \cellcolor{yellow}\textbf{0.0409} & \cellcolor{yellow}\textbf{0.0159} & \cellcolor{yellow}\textbf{0.0395} & \cellcolor{yellow}\textbf{424.46} & \cellcolor{yellow}0.197 & \cellcolor{yellow}\textbf{43.94} & 
 \cellcolor{yellow} 54.96 \\
\bottomrule
\end{tabular}
}
\end{table*}

\subsection{Experimental Setup}
\label{subsec:Setup}

\subsubsection{Implementation Details} 

\noindent In our experiments, the resolution of the input speaker image sequence is set as $224 \times 224$, based on which the ReactFace model is trained with an AdamW optimizer \cite{kingma2014adam} with a fixed learning rate of $2e^{-5}$, $\beta_1 = 0.9$ and $\beta_2 = 0.999$. Our minibatch size is set to 4.
For hyper-parameters balancing loss terms in Sec.~\ref{subsec:training}, we set  $\lambda_{\text{kl}} = 1e^{-5}$, $\lambda_{\text{smo}} = 10$, and $\lambda_{\text{div}} = 100$. The momentum parameter $\alpha$ in Sec.~\ref{subsec:Multiple} is empirically set to 0.999. Our code is implemented in PyTorch \cite{paszke2019pytorch} using a single Tesla A100 GPU with 80G memory and runs for total 200 epochs. 

To extract 3D expression and pose coefficients from videos, we leverage the state-of-the-art 3DMM models FaceVerse \cite{wang2022faceverse}, which estimates the pose, expression, shape, texture and lighting parameters, and defines expression coefficients $\psi$ ($\lvert \psi \rvert = 52$) along with pose coefficients $\theta$ (\emph{i.e.,} 3-dimensional translation and the 3-dimensional rotation), resulting in totally 58 coefficients for each face frame. The reason for choosing FaceVerse 3DMM system \cite{wang2022faceverse} is that each of its expression coefficient corresponds to an ARKit blendshape, which has an explicitly and human interpretable definition such as 'browInnerUp', 'eyeLookDownRight ', ' jawOpen', ' mouthFunnel', 'noseSneerRight' and 'tongueOut'. In contrast to the principal component analysis (PCA) blendshapes employed in conventional 3D Morphable Model (3DMM) systems, the ARKit face blendshapes \footnote{https://arkit-face-blendshapes.com} used in FaceVerse constitute discrete micro-expression components. These components afford a finer-grained representation of facial muscle movements, enabling a more precise and nuanced characterization of facial expressions. In short, the use of ARKit blendshapes enhances the degree of controllability in the process of facial expression generation. Both expression and pose coefficients are then used for image rendering. 
In this paper, we fine-tune the PIRender \cite{ren2021pirenderer} to fit the target 3DMM system and utilize it to translate predicted 3DMM coefficients to a portrait of the listener.

\subsubsection{Baselines}

We leverage five baselines (\emph{e.g.,}  Mirror, Random, NN motion, NN audio and Trans-AE) and reproduce three recently proposed facial reaction generation methods (\emph{e.g.,} LFI~\cite{jonell2020let}, Ng et al. \cite{ng2022learning} and Zhou et al. \cite{zhou2022responsive}) as the baseline facial reaction generation models, where we strictly followed the training settings given in the corresponding papers:
\begin{itemize}

    \item \textbf{Mirror:} Mirror refers to replicating the facial movements of the speaker.

    \item \textbf{Random:} Random involves sampling reactions from Gaussian distributions.

     \item \textbf{NN motion:} NN motion searches for the nearest neighbor (NN) of the current speaker motion segment and returns the corresponding listener segment, a method commonly used in graphics synthesis.

      \item \textbf{NN audio:} NN audio searches for the NN through the speaker's auditory signals.
    
    \item \textbf{Trans-AE:} An autoencoder model with transformer architecture, which has the same speaker behaviour encoder and alignment module as that in our ReactFace but a straightforward decoder outputting reactions. 

        \item \textbf{LFT~\cite{jonell2020let}:} a flow-based generative model that has demonstrated strong performance in generating listener head motions.

    \item \textbf{Ng et al. \cite{ng2022learning}:} A VQ-VAE \cite{van2017neural} based model with a motion-audio
cross attention transformer branch encoding  speaker's 3DMM coefficients and audio signals.

    \item \textbf{Zhou et al. \cite{zhou2022responsive}:} a simple LSTM model which takes speaker's 3DMM coefficients, diverse audio features and initial listener 3DMM coefficients at first frame as input and outputs listener 3DMM coefficients in a sequence-to-sequence manner.
    
\end{itemize}
To ensure a fair and comprehensive comparison, we additional present our results on the RLD dataset, which has been employed in Zhou et al. \cite{zhou2022responsive}. To produce final facial reactions, we followed the same protocol as our approach, including the utilization of the identical 3D Morphable Model (3DMM) system \cite{blanz2023morphable} as employed in \cite{zhou2022responsive}.

\subsubsection{Evaluation Metrics} 

In this paper, we follow \cite{song2023multiple} to evaluate four aspects (\emph{e.g.,} appropriateness, diversity, realism, and synchrony) of the generated facial reactions, and also assess the generation speed. Firstly, we use the Facial Reaction Correlation (\textbf{FRCorr}) metric proposed in to assess the \textbf{appropriateness} by measuring the DTW distance and correlations between the generated facial reactions and their most similar appropriate real facial reactions.

Then, diversity is evaluated by (1)
inter-condition diversity \textbf{FRDvs}: the average MSE distance between all pairs of generated reactions conditioned on different speaker behaviours ($N$ different speaker sequences) to measure diversity across speaker behaviour conditions, which is computed as:
\begin{equation}
\text{FRDvs} = \frac{\sum^{M}_{j=1}\sum^{N-1}_{i=1} \sum^{N}_{k = i+1}
\text{MSE}(\bm{\hat{\mathrm{R}}}^{l}_{i, j, 1:T}, \bm{\hat{\mathrm{R}}}^{l}_{k, j, 1:T})}{N(N-1)M};    
\end{equation}
where $N$, $M$, and $T$ denotes the number of sequence pairs, sampling times for diverse facial reaction generation and the length of a video clip, respectively. 
(2) inter-sample diversity \textbf{FRDiv}: the average $L2$ distance between all pairs of generated reactions in response to the same speaker behaviour, which is computed as:
\begin{equation}
\text{FRDiv} = \frac{\sum^{N}_{i=1} \sum^{M-1}_{j=1} \sum^{M}_{k=j+1} \| \bm{\hat{\mathrm{R}}}^{l}_{i, j, 1:T}  - \bm{\hat{\mathrm{R}}}^{l}_{i,k,1:T} \|_2^{2}}{NM(M-1)};    
\end{equation}
and (3) inter-frame diversity \textbf{FRVar}: average 
variation of generated reaction frames within a clip ($T$ frames) to measure diversity within one generated facial reaction sequence, which is computed as: 
\begin{equation}
\text{FRVar} = \frac{1}{N \times M}\sum^{N}_{i=1}\sum^{M}_{j = 1} \text{var}(\bm{\hat{\mathrm{R}}}^{l}_{i, j, 1:T});
\end{equation}

Thirdly, realism \textbf{FRRea} is measured using 
Fr'{e}chet Video Distance (FVD) \cite{unterthiner2018towards} metric. Finally, we employ the Time Lagged Cross Correlation (TLCC) metric to assess the \textbf{synchrony (FRSyn)} between speaker behaviours and the corresponding generated facial reactions. Please refer to \cite{song2023multiple} for the more details of these metrics. We utilize Frames Per Second (FPS) as a key metric to assess the generation speed of each generative model.

\subsection{Qualitative Results}
\label{subsec:Qualitative Results}

We present a qualitative comparison in Fig.~\ref{fig:vis_diff_sample} to visually compare the appropriateness, diversity, realism, and synchrony of the facial reactions generated by our ReactFace model with two state-of-the-art methods \cite{ng2022learning,zhou2022responsive}. The result shows that facial reactions generated by our approach have \textbf{higher diversity} and are better synchronised with the speaker behaviour compared to competitors that can only produce perturbations based on a single facial expression. For example, at time $t=601$, our ReactFace generated a smiling facial reaction that corresponds to the happy facial expression of the speaker. Also, both speaker facial expression and the generated facial reaction are transferred to different states at the time $t = 701$. In contrast, the baselines produced almost identical facial expressions throughout the sequence, and not synchronised with speaker behaviours. In summary, our ReactFace produced appropriate, diverse, continuous and synchronised facial reactions, which are superior to baselines.

\subsection{Quantitative Results}
\label{subsec:Quantitative Results}

We report the quantitative results in Table~\ref{tb:compare_hb}. It is clear that the proposed ReactFace generated facial reactions with the best diversity and realism and synchrony among all competitors (\emph{i.e.,} recently proposed state-of-the-art facial reaction generation approaches) \cite{ng2022learning, zhou2022responsive, jonell2020let} as well as the baseline Trans-AE (\emph{i.e.,} a transformer-based AutoEncoder that has the same architecture of our speaker behaviour encoding network, which directly maps the speaker behaviour features to facial reaction predictions). We explain such results as the training process of these competitors only pairs a single real facial reaction label with the input speaker behaviour, which lead to their models being optimised to generate the average facial behaviour of multiple appropriate facial reactions in response to each speaker behaviour, \emph{i.e.,} this results in suboptimal diversity and appropriateness of the generated facial reactions. Also, the novel SLBS module also ensures the ReactFace to always generate facial reactions synchronised with speaker behaviours, making our ReactFace largely outperform competitors in terms of the synchrony. More importantly, we also found that the facial reactions generated by ReactFace are also appropriate, \emph{i.e.,} it achieved competitive results in term of appropriateness metric with 0.197 FRCorr, which is ranked at the second among all compared approaches. Although Trans-AE achieved the best appropriateness, it sacrifices the diversity of generated results, with 0.0063 FRDvs and 0.0003 FRVar.

Additionally, our ReactFace model achieves an impressive performance of over 54.96 FPS on a single Tesla A100 GPU, making it highly suitable for real-time interactive scenarios that require a minimum of 30 FPS. It is important to note that ReactFace handles raw audio and video frame processing directly, without the need for pre-processing steps such as mapping video frames to 3DMM coefficients. 
In contrast, the approaches by Zhou et al.~\cite{zhou2022responsive}, Ng et al.~\cite{ng2022learning}, LFI~\cite{jonell2020let}, and NN motion require an additional pre-processing step to extract 3DMM coefficients from video frames.
The FPS evaluation also accounts for the entire Facial Reaction Visualization process.

\subsection{Ablation Studies}
\label{subsec:ablation}

In this section, we provide a set of ablation study results to systematically investigate: (i) contributions of different modalities; (ii) effectiveness of the proposed components in AFRG and SLBS modules; and (iii) contributions of different losses. In addition, Fig. \ref{fig:vis_attn_weights} further provides visualisation results to explain the temporal alignments achieved by our proposed MIM and VIM. \\

\noindent \textbf{Contributions of different modalities:} Table~\ref{tb:compare_ablation_study_modalities} evaluates the importance of different modality contributing to the final generated facial reactions. Our analysis yielded three key findings: (i) both speaker facial and speech behaviours are important in evoking listener's facial reactions; (ii) excluding facial modality resulted in low synchrony of the generated facial reactions; and (iii) in our framework, speaker speech behaviours made more contributions than facial behaviours in generating more diverse, realistic, and appropriate facial reactions. \\

\noindent \textbf{Effectiveness of MIM and VIM in the SLBS module:} Table~\ref{tb:compare_ablation_study_comnponents} evaluates the importance of different components of our ReactFace. The results show that MIM (interaction between speaker audio signals and listener reactions) made large contributions to the Diversity (FRDiv) and Appropriateness (FRCorr) of the generated facial reactions. Meanwhile, our VIM (interaction between speaker facial behaviour and listener reactions) plays a key role in improving the diversity, realism, and appropriateness performances. While employing either VIM or MIM improves the performance, combining the two leads to substantial gains in synchrony, appropriateness, and diversity of generated reactions. This improvement may come from enriched interaction features derived from multiple perspectives, including both speaker's verbal and non-verbal cues. \\

\noindent \textbf{Comparison between MIM in the SLBS module and p2p audio-visual alignment \cite{fan2022faceformer}:} The comparative analysis presented in Table~\ref{tb:compare_ablation_study_MIM_p2p} shows different results achieved by the modality interaction module (MIM) and the p2p audio-visual alignment method used in Faceformer \cite{fan2022faceformer} performed. These results clearly show that MIM outperforms the p2p method in many aspects. Specifically, the p2p audio-visual alignment technique that has been leveraged to synchronize visual lip movements with human audio at individual timestamps lacks consideration for comprehensive conversation context, leading to poorer appropriateness (FRCorr) and less varied reaction patterns (in terms of FRVar and FRDiv) and reaction realism (FRRea). \\

\noindent \textbf{Effectiveness of the momentum employed in the AFRG module (in sampling block):} As shown in Table~\ref{tb:compare_sampling},  it is clear that momentum operation is essential for the sampling process, ensuring the appropriateness (FRCorr) and realism (FRRea) of the generated facial reactions. The reason is that the absence of momentum in reaction sampling leads to an abrupt transition between the previous and current frames' windows, making the generated facial reaction videos lacking the temporal continuity inherent in human behaviour. While momentum operation reduces the facial reaction diversity (FRDiv), it prevents abrupt changes and jitters, ensuring a more realistic and stable video output. \\

\noindent \textbf{Contributions of different losses:} Table~\ref{tb:compare_ablation_loss_function} then demonstrates the contributions of the loss functions employed in our training. Our observations are presented as follows: (i) adding a 3D speaker face reconstruction loss $\mathcal{L}_{rec}^s$ allows MSBEA to extract semantic meaningful features from speaker facial behaviours, which results in improved realism, appropriateness and synchrony of the generated facial reactions; (ii) Kullback-Leibler divergence loss $\mathcal{L}_\text{kl}$ and energy-based diversity loss $\mathcal{L}_\text{div}$ are critical in generating diverse facial reactions, while $\mathcal{L}_\text{div}$ additionally facilitates our ReactFace to learn different appropriate facial reactions in response to each speaker behaviour; and (iii) the smooth loss $\mathcal{L}_{smo}$ enhancing the appropriateness (in terms of the FRDist) of the generated facial reactions by smoothing variations between adjacent frames and avoiding jitter.

\begin{table}[t]
 \caption{
Ablation study results achieved for different components.} 
 \resizebox{0.48\textwidth}{!}{
\begin{tabular}{cc|ccccc}
\toprule
Speech & Face   &\textbf{FRVar} & \textbf{FRDiv} & \textbf{FRRea}  & \textbf{FRCorr}  & \textbf{FRSyn} \\ \midrule
  \checkmark    &             &0.0173  &0.0351  &450.33 & 0.168 &46.33  \\ 
  
      &  \checkmark           &0.0155   &0.0217  &597.35   &0.123 &44.38  \\  

 \checkmark  &  \checkmark           &0.0159  &0.0395 &424.46  &0.197  &43.94  \\
\bottomrule
\end{tabular}
}
\label{tb:compare_ablation_study_modalities} 
\end{table}

\begin{table}[t]
 \caption{
Ablation study results achieved for different components.} 
 \resizebox{0.48\textwidth}{!}{
\begin{tabular}{cc|ccccc}
\toprule
MIM & VIM   &\textbf{FRVar} & \textbf{FRDiv} & \textbf{FRRea}  & \textbf{FRCorr}  & \textbf{FRSyn} \\ \midrule

     &             &0.1824  &0.0001  &526.09  &0.102 &45.42  \\  

  \checkmark    &             &0.0107  &0.0180  & 668.38 &0.123 &46.30  \\

      &  \checkmark           &0.0190  &0.0315  &460.75   &0.184 &45.36  \\

 \checkmark  &  \checkmark          &0.0159  &0.0395 &424.46 &0.197  &43.94 \\
\bottomrule
\end{tabular}
}
\label{tb:compare_ablation_study_comnponents} 
\end{table}

\begin{table}[t]
 \caption{
Ablation study results achieved for different audio-visual alignment strategies.} 
 \resizebox{0.48\textwidth}{!}{
\begin{tabular}{c|ccccc}
\toprule
  &\textbf{FRVar} & \textbf{FRDiv} & \textbf{FRRea}  & \textbf{FRCorr}  & \textbf{FRSyn} \\ \midrule

  P2P audio-visual alignment     &0.0129  &0.0321  & 594.14 &0.185 &44.40  \\ 
 MIM    &0.0159  &0.0395 &424.46 &0.197  &43.94   \\  
\bottomrule
\end{tabular}
}
\label{tb:compare_ablation_study_MIM_p2p} 
\end{table}

\begin{table}[t]
 \caption{
Ablation study results achieved for different audio-visual alignment strategies.} 
 \resizebox{0.48\textwidth}{!}{
\begin{tabular}{c|ccccc}
\toprule
  &\textbf{FRVar} & \textbf{FRDiv} & \textbf{FRRea}  & \textbf{FRCorr}  & \textbf{FRSyn} \\ \midrule

  sampling w/o momentum    &0.0035  &0.1002  & 614.36  &0.146 &45.56  \\ 
 sampling w/ momentum    &0.0159  &0.0395 &424.46 &0.197  &43.94   \\  
\bottomrule
\end{tabular}
}
\label{tb:compare_sampling} 
\end{table}

\begin{table}[t]
 \caption{
 Ablation study
 results achieved for applying different loss functions at the training stage.} 
 \resizebox{0.48\textwidth}{!}{
\begin{tabular}{ccccc|ccccc}
\toprule
$\mathcal{L}_{rec}^a$ & $\mathcal{L}_{rec}^s$ & $\mathcal{L}_{kl}$ &$\mathcal{L}_{div}$ & $\mathcal{L}_{smo}$  & \textbf{FRVar} & \textbf{FRDiv}  & \textbf{FRRea}   & \textbf{FRCorr} & \textbf{FRSyn} \\ \midrule
   &  \checkmark    &  \checkmark       &  \checkmark    &  \checkmark  &0.0220
     & 0.0433  &412.14    &0.166    &45.96\\

    \checkmark &  &  \checkmark     &  \checkmark    & \checkmark &
    0.0054
     &0.0831   & 596.22   & 0.143    &46.26\\ 
     
  \checkmark    & \checkmark   &        &  \checkmark  &  \checkmark  &0.0156
     &0.0238   &479.77    & 0.186   &45.36   \\

    \checkmark   & \checkmark  &  \checkmark       &    & \checkmark &
    0.0001
     &0.0000   & 590.24    & 0.112    &43.39 \\  
     
           \checkmark   & \checkmark  &  \checkmark       & \checkmark    &
     &0.0171   &0.0248    &420.61  &0.139    &44.16 \\
     
     \checkmark   & \checkmark  &  \checkmark       &  \checkmark    &  \checkmark    &0.0159   &0.0395 &424.46 &0.197  &43.94 \\
\bottomrule
\end{tabular}
}
\label{tb:compare_ablation_loss_function} 
\end{table}

\subsection{Visualization Analysis}

\begin{figure}[t!]
    \centering
    \includegraphics[width=1\columnwidth]{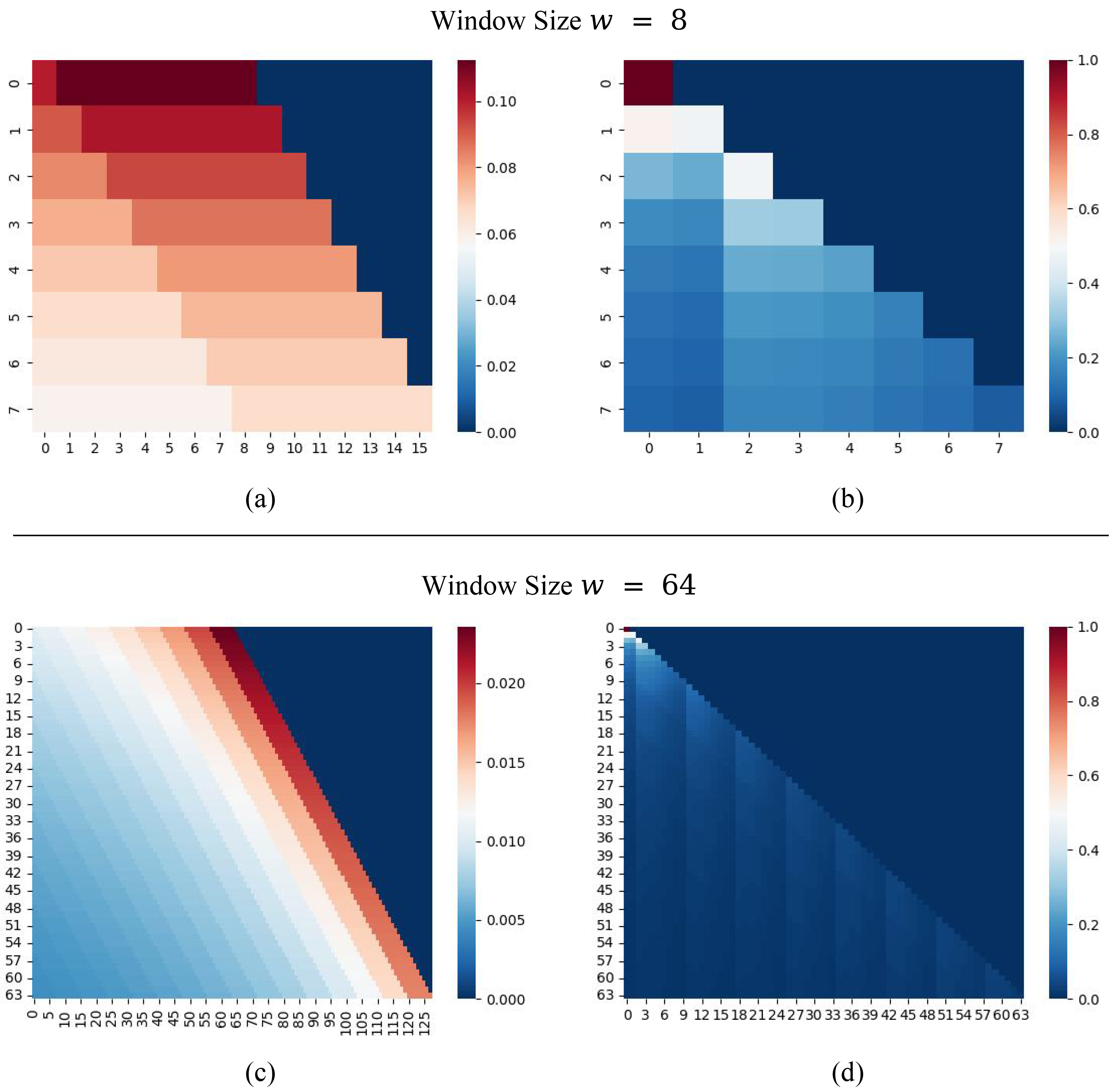} 
    \caption{Visualizations of attention weights in MIM and VIM. Attention weights of the (a) modality interaction module and (b) visual interaction module of the decoder, when setting window size $w = 8$ at the inference stage. Attention weights of the (c) modality interaction module and (d) visual interaction module of the decoder, when setting window size $w = 64$ at the training stage. }
    \label{fig:vis_attn_weights}
\end{figure}

\noindent To provide more insightful explanation of the proposed two speaker-listener interaction modules (\emph{i.e.,} MIM and VIM), Fig.~\ref{fig:vis_attn_weights} visualises the attention weights within the transformer decoder layers of these modules. Specifically, we calculate the average attention weights across all attention heads, with each representing the correlation between a pair of frames. For the MIM, Fig.~\ref{fig:vis_attn_weights}(a) and Fig.~\ref{fig:vis_attn_weights}(c) correspond to the correlation between speaker speech frames and the generated facial reaction frames, while for VIM, Fig.~\ref{fig:vis_attn_weights}(b) and Fig.~\ref{fig:vis_attn_weights}(d) represent the correlation between speaker facial frames and the generated facial reaction frames. We perform this visualisation at both the inference stage (Fig.~\ref{fig:vis_attn_weights}(a) and Fig.~\ref{fig:vis_attn_weights}(b)) and the training stage (Fig.~\ref{fig:vis_attn_weights}(c) and Fig.~\ref{fig:vis_attn_weights}(d)), with window sizes set to 8 and 64, respectively. As illustrated in Fig.~\ref{fig:vis_attn_weights}, several key observations can be found: (i) later frames have zero influences on earlier facial reaction frames' generation, preventing the generation of existing facial reaction frames to be influenced by future speaker behaviours; (ii) the attention mechanism establishes both long-range and short-range context dependencies between the speaker speech/visual feature frames and facial reactions during the reasoning, emphasising the importance of contextual information; and (iii) Notably, a facial reaction frame tends to prioritize nearby speaker auditory/visual frames in its attention mechanism, while attenuating its focus on more distant speaker behaviour frames.

\subsection{Perception Survey}
\label{subsec:perceptual}

\noindent We conducted user studies on the Tencent Questionnaire platform to evaluate the facial reactions generated by our ReactFace in comparison to these generated three competitors:  Ng et al \cite{ng2022learning}, Zhou et al. \cite{zhou2022responsive} and the GT real facial reactions. Specifically, a total of 41 volunteers from Shenzhen University are recruited, comprising 12 females and 29 males, aged between 22 and 45. These volunteers, primarily from the college of computer science and software engineering and possessing relevant machine learning or deep learning expertise, were instructed with an online survey. This survey aimed to determine their preferences (\emph{i.e.,} choosing a video containing more appropriate, realistic and synchronised facial reaction from a pair of compared videos) between the facial reaction videos generated by ReactFace and competitors. Specifically, In total, each volunteer watched 8 video clips (24 video pairs), with each video clip containing three pairs of videos, \emph{i.e.,} comparing our ReactFace with Ng et al \cite{ng2022learning}, Zhou et al. \cite{zhou2022responsive} and the GT real facial reactions, respectively. The platform displayed randomized video pairs for each comparison, and volunteers were instructed to evaluate the quality of the generated facial reactions in terms of realism, diversity, appropriateness, and synchronization. Results displayed in Tab.~\ref{ex:user_study} indicate that the facial reactions generated by our ReactFace were generally favored by more than 80\% human participants over the competitor in all cases. Interestingly, ReactFace was even preferred over ground truth samples in 17.07\% and 19.51\% of the evaluated cases in terms of appropriateness and realism, respectively.

\begin{table}[t!]
\centering
 \caption{\label{tb:user-study} User preference results between the facial reactions generated by our ReactFace and competitors.
}
 \resizebox{0.48\textwidth}{!}{
 \begin{tabular}{lccccc}
 \toprule
\multicolumn{2}{l|}{\textbf{Ours \emph{vs.} Competitor}}&\textbf{Realism}  &\textbf{Diversity} &\textbf{Appropriateness} &\textbf{Sync}   \\
\midrule 
\multicolumn{2}{l|}{Ours \emph{vs.} Ng et al. \cite{ng2022learning}}& 90.24\%  & 95.12\%  &  87.80\% &  92.68\%  \\
\multicolumn{2}{l|}{Ours \emph{vs.} Zhou et al. \cite{zhou2022responsive}}&80.49\%  & 90.24\%  & 82.93\%   &87.80\%  \\
\multicolumn{2}{l|}{Ours \emph{vs.} GT}& 19.51\%   &9.76\%  &17.07\%  &14.63\%   \\
\bottomrule
 \end{tabular} 
 }
\label{ex:user_study}
\end{table}

\subsection{Additional Results on RLD Dataset}

\begin{table}[t!]
\centering
 \caption{\label{tb:compare_RLD} 
Comparison of quantitative results on RLD dataset.} 
\resizebox{0.45\textwidth}{!}{
 \begin{tabular}{lccccc}
 \toprule
\multicolumn{2}{l|}{\textbf{Methods}} & \textbf{FRRea}   & \textbf{FRVar}  & \textbf{FRDiv} & \textbf{FRSyn}\\
\midrule 

\multicolumn{2}{l|}{GT} & 168.24 

 &1.8439 & -
&29.61
 \\

\multicolumn{2}{l|}{ Trans-AE} & 250.09

 & 0.0145  &  0
& 32.52
 \\
\multicolumn{2}{l|}{ Ng et al. \cite{ng2022learning}} & 460.48

 &\textbf{1.1032}  &   0
& \textbf{31.00}
 \\
\multicolumn{2}{l|}{Zhou et al. \cite{zhou2022responsive}} &  \textbf{180.56}
&   
 0.9314 & 0& 32.62   \\

\multicolumn{2}{l|}{\cellcolor{yellow}ReactFace w/o $\mathcal{L}^a_{rec}$}&\cellcolor{yellow}271.09
 &\cellcolor{yellow}0.3539  &\cellcolor{yellow}\textbf{0.3015} &\cellcolor{yellow}31.12\\
\bottomrule
 \end{tabular} 
 }
\end{table}
\noindent In this section, we extend our experimental analysis to further include results on the RLD dataset \cite{zhou2022responsive}. This dataset comprises data from 92 subjects, consisting of 67 speakers and 76 listeners, with a total of 483 video clips sourced from YouTube. Notably, the RLD dataset lacks 'appropriate facial reaction' labels. Consequently, we present results obtained using our ReactFace model, trained without the utilization of the appropriate facial reaction reconstruction loss $\mathcal{L}^a_{rec}$. In this particular configuration, we replace $\mathcal{L}^a_{rec}$ with a standard reconstruction loss $\mathcal{L}_{rec}$, which solely assesses the dissimilarity between each generated facial reaction and its corresponding Ground Truth (GT) facial reaction. It's important to acknowledge that the limited scale of the RLD dataset, comprising 483 video clips, presents challenges for training models based on Transformer architecture. This is because Transformer-based models typically require more data for optimization as compared to LSTMs \cite{hochreiter1997long}, and CNNs \cite{krizhevsky2012imagenet, lecun1998gradient, he2016deep}. Consequently, both our ReactFace and the Trans-AE models struggle in this scenario. However, despite these challenges, as demonstrated in Table~\ref{tb:compare_RLD}, our ReactFace model manages to achieve competitive performance in terms of realism, diversity, and synchrony when compared to state-of-the-art approaches.

\subsection{Failure Cases}
\label{sec:failure_cases}

\begin{figure}[t]
    \centering
    \includegraphics[width=0.9\columnwidth]{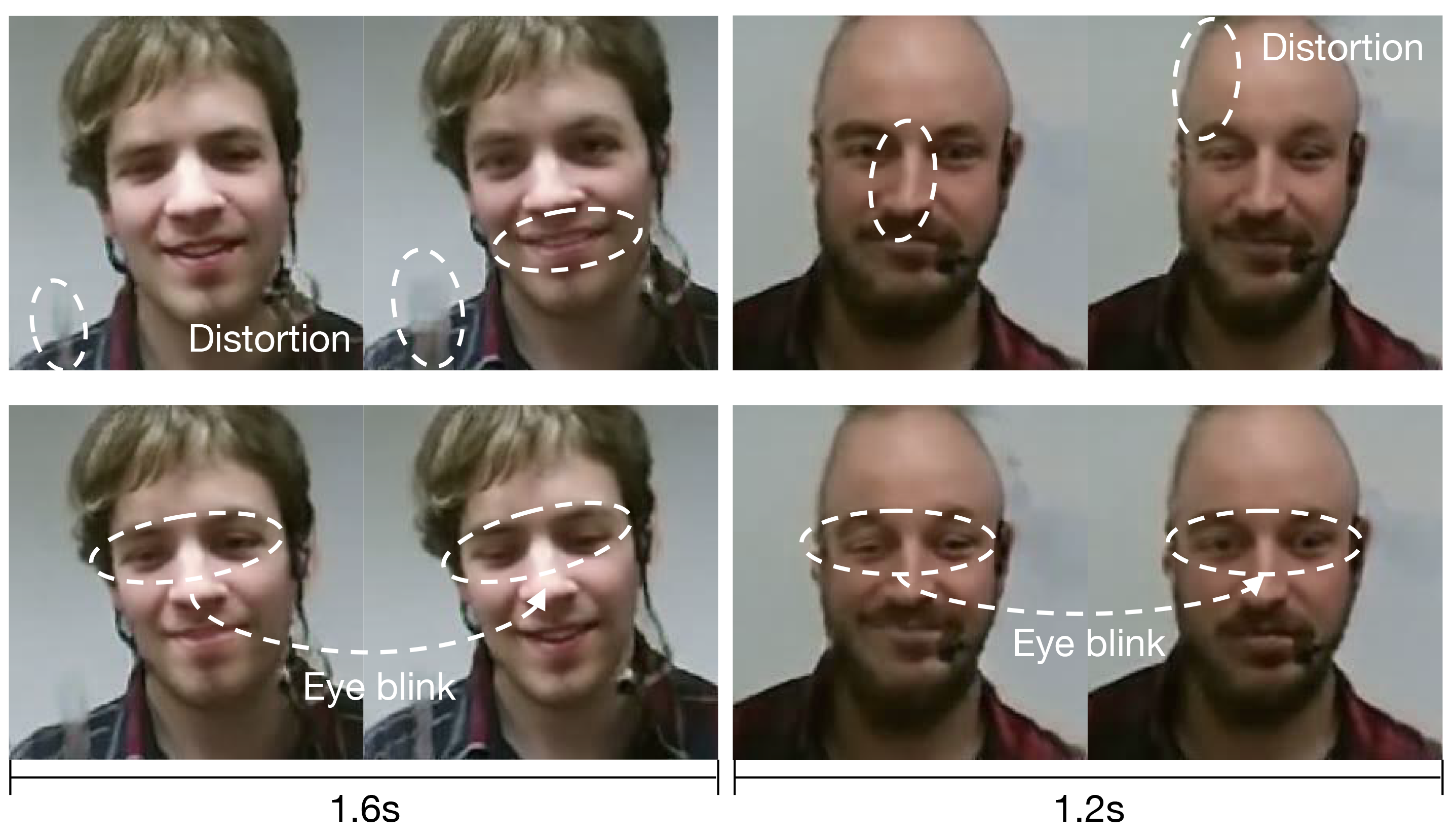} 
    \caption{Failure cases of generated listener facial reaction by ReactFace.} 
    \label{fig:failure_cases}
\end{figure}

\noindent In this section, we discuss several failure cases observed in our generated samples.
Fig.~\ref{fig:failure_cases} presents the failure cases of our method on REACT2023 dataset. The qualitative results indicate that these failures primarily fall into two categories: (1) the rendering network for facial reaction visualisation may introduce image distortions, affecting both the face and background, as illustrated in the first row of Fig.~\ref{fig:failure_cases}; (2) ReactFace struggles to capture the dynamics of eye blinking, resulting in a slower blink speed (over one second) compared to natural human blinks, which typically occur within 100–150 milliseconds.

\section{Conclusion}

This paper proposes the first online MAFRG model (called ReactFace) capable of generating multiple appropriate facial reactions in response to each speaker behaviour. Our ReactFace starts with processing the input speaker behaviour using a multi-modal speaker behaviour encoding and alignment (MSBEA) module, and then produces multiple facial reactions based on an appropriate facial reaction generation (AFRG) module, where a speaker-listener behaviour synchronisation (SLBS) module is introduced to model interactions between speaker behaviours and the corresponding generated facial reactions (\emph{i.e.,} synchronising speaker behaviours and the corresponding generated facial reactions).

We systematically assess the efficacy of our ReactFace through a rigorous evaluation conducted on the combination of two dyadic interaction datasets provided by an open appropriate facial reaction generation challenge \cite{song2023react2023}, leveraging various metrics encompassing diversity, realism, appropriateness, and synchrony that also have been employed by the challenge \cite{song2023react2023}. The obtained results demonstrate the effectiveness of our multiple facial reaction generation strategy. This strategy enables our model to produce a spectrum of diverse and contextually appropriate facial reactions. Furthermore, our introduced synchronization strategy significantly contributes to enhancing the realism of the generated reactions. Note that these reactions exhibit a high degree of continuity and synchrony with the corresponding audio-visual behaviours of the speaker. In-depth ablation analysis of the individual modules in our framework further demonstrates their effectiveness and indispensability. These modules, meticulously designed and integrated, play pivotal roles in augmenting the overall performance and cohesiveness of the proposed system.

Nevertheless, given the pioneering nature of our approach in the field of multiple appropriate facial reaction generation, certain modules employed may not yet represent the optimal configurations, leaving room for refinement in future investigations. Furthermore, there is a pressing need for the model to undergo further refinement to ensure its efficiency and practicality when applied in real-world scenarios. Finally, due to the limited resources, we were unable to reproduce all existing generative methods for comparison. Thus, our future work will focus on not only exploring more effective and efficient algorithms for online MAFRG task but also benchmarking existing generative methods (\emph{e.g.,} GANs and diffusion models) for both offline and online MAFRG tasks.

\ifCLASSOPTIONcompsoc
  \section*{Acknowledgments}
\else
  \section*{Acknowledgment}
\fi
C. Luo, X. Xie, and L. Shen are supported by the Natural Science Foundation of China under grants no. 62276170, 82261138629, 
the Science and Technology Project of Guangdong Province under grants no. 2023A1515011549, 2023A1515010688, Guangdong Provincial Key Laboratory under Grant 2023B1212060076, and
the Science and Technology Innovation Commission of Shenzhen under grant no. JCYJ20220531101412030.
S. Song is partially supported by the European Union’s Horizon 2020 Research and Innovation programme under grant agreement No. 826232. H. Gunes and M. Spitale are supported by the EPSRC under grant ref. EP/R030782/1.
Z. Ge is supported by National Health and Medical Research Council APP2006551 and NHMRC APP2009923.

\bibliography{egbib}

\bibliographystyle{IEEEtran}

\vfill

\end{document}